\documentclass[10pt]{article} 

\usepackage[preprint]{assets/rlj}           

\usepackage{amssymb}            
\usepackage{mathtools}          
\usepackage{mathrsfs}           
\usepackage{graphicx}           
\usepackage{subcaption}         
\usepackage[space]{grffile}     
\usepackage{url}                
\usepackage{lipsum}             

\usepackage{float}
\usepackage{xcolor}
\usepackage{booktabs}
\usepackage{placeins}
\usepackage{comment}
\usepackage[noabbrev,capitalize]{cleveref}
\hypersetup{
  colorlinks   = true,              
  urlcolor     = blue,              
  linkcolor    = blue,              
  citecolor    = teal                
}

\newcommand{\mytitle}{Preventing Learning Stagnation in PPO by \\Scaling to 1 Million Parallel Environments}
\newcommand{\myrunningtitle}{Preventing Learning Stagnation in PPO by Scaling to 1 Million Parallel Environments}
\title{\mytitle}

\setrunningtitle{\myrunningtitle}

\author{Michael Beukman\textsuperscript{1,$*$}, Khimya Khetarpal\textsuperscript{2}, Zeyu Zheng\textsuperscript{2}, \\Will Dabney\textsuperscript{2}, Jakob Foerster\textsuperscript{1}, Michael Dennis\textsuperscript{2}, Clare Lyle\textsuperscript{2}}

\emails{\{mbeukman,jakob\}@robots.ox.ac.uk\\\{khimya,zeyuzheng,wdabney,dennismi,clarelyle\}@google.com}

\affiliations{
$^{1}$\textbf{FLAIR, University of Oxford}\\
$^{2}$\textbf{Google DeepMind}
\par 
$^*$ Work done while at Google DeepMind.
}

\contribution{
    }
    {
    }

\contribution{
    }
    {
    }

\contribution{
    }
    {
    }


\summary{
An agent's performance stagnating at a suboptimal level is a common problem in deep on-policy RL.
Focusing on PPO, we show that plateaus in certain regimes arise not because of known exploration, capacity, or optimisation challenges, but because sample-based estimates of the loss eventually become poor proxies for the true objective over the course of training. 
Looking deeper, PPO alternates between sampling rollouts from several parallel environments \textit{online} using the current policy (which we call the ``outer loop'') and performing repeated minibatch SGD steps against this \textit{offline} dataset (the ``inner loop''). 
In our work, we abstract away the inner loop, and conceptually model the outer loop as standard stochastic optimisation. The step size is then controlled by the regularisation strength towards the previous policy and the gradient noise by the number of samples collected between policy update steps. 
This framing predicts that, much like in SGD, if the outer step size is too large relative to the noise, updates become uninformative and lead to the policy thrashing around a local optimum instead of converging.
Recasting PPO in this light makes it clear that there are two ways to address this particular type of learning stagnation: either reduce the step size or increase the number of samples collected between updates. 
We validate the predictions of our model and conclude that increasing the number of parallel environments is a simple way to avoid these plateaus by simultaneously altering both these factors. 
Applying our analysis and scaling PPO to more than 1M parallel environments enables monotonic performance improvement up to one trillion transitions and leads to vastly superior performance compared to prior baselines in a complex open-ended domain.

}

\begin{document}
\maketitle  

\begin{abstract}
An agent's performance stagnating at a suboptimal level is a common problem in deep on-policy RL.
Focusing on PPO, we show that plateaus in certain regimes arise not because of known exploration, capacity, or optimisation challenges, but because sample-based estimates of the loss eventually become poor proxies for the true objective over the course of training. 
Looking deeper, PPO alternates between sampling rollouts from several parallel environments \textit{online} using the current policy (which we call the ``outer loop'') and performing repeated minibatch SGD steps against this \textit{offline} dataset (the ``inner loop''). 
In our work, we abstract away the inner loop, and conceptually model the outer loop as standard stochastic optimisation. The step size is then controlled by the regularisation strength towards the previous policy and the gradient noise by the number of samples collected between policy update steps. 
This framing predicts that, much like in SGD, if the outer step size is too large relative to the noise, updates become uninformative and lead to the policy thrashing around a local optimum instead of converging.
Recasting PPO in this light makes it clear that there are two ways to address this particular type of learning stagnation: either reduce the step size or increase the number of samples collected between updates. 
We validate the predictions of our model and conclude that increasing the number of parallel environments is a simple way to avoid these plateaus by simultaneously altering both these factors. 
Applying our analysis and scaling PPO to more than 1M parallel environments enables monotonic performance improvement up to one trillion transitions and leads to vastly superior performance compared to prior baselines in a complex open-ended domain.

\end{abstract}
\section{Introduction}
A common failure mode in RL is the tendency for an agent's performance to plateau well below the theoretical optimal return in an environment~\citep{nikishin2022primacy,lyle2022understanding,nauman2024bigger}. 
This is becoming an increasingly visible problem as highly-parallelised and complex RL environments have gained popularity~\citep{brax2021Github,makoviychuk2021Isaac,gymnax2022github,nikulin2023xlandminigrid,rutherford2023jaxmarl,matthews2024Craftax,mujoco_playground_2025}, meaning that it is becoming feasible to run agents for billions or trillions of timesteps with only modest hardware requirements~\citep{matthews2024kinetix}. 
However, if our algorithms cannot improve beyond a subpar plateau even in the limit of additional experience, there is little use for these trillions of timesteps.

Prior work has explored different reasons behind why an algorithm might plateau. One explanation is plasticity loss or the primacy bias, where the network accumulates pathologies during training that hinder the optimisation process~\citep{nikishin2022primacy,lyle2022understanding,lyle2024disentangling}.
Other works highlight insufficient exploration~\citep{thrun1992efficient,bellemare2016unifying,kuttler2020Nethack,taiga2021bonus}, e.g., due to the agent collapsing to a near-deterministic policy too early. While this may be a problem in certain settings, empirical results suggest that plateaus can still occur in dense reward tasks which do not pose hard exploration challenges, highlighting the importance of implementation details and hyperparameters~\citep{henderson2018deep,engstrom2020implementation,andrychowicz2020matters}. 

\looseness=-1 We take a different perspective, one inspired by PPO's roots in proximal gradient methods and the empirical similarities between plateaus in RL and stochastic optimisation. In particular, we abstract away the inner neural network optimisation process and focus only on the outer loop, conceptually modelling it as standard stochastic optimisation. In this model, the \textit{step size} represents how much the policy changes between update iterations, whereas the \textit{update noise} represents how well minimising the loss on a sampled batch of trajectories corresponds to maximising the true objective.
It now becomes clear that PPO is vulnerable to plateaus when the outer step size is too large relative to the update noise level; much like in SGD, this prevents convergence by causing the policy to thrash around a local optimum. Crucially, such a mechanism suggests that these plateaus are caused not by the policy changing slowly, but by it changing too much between update iterations.
There are therefore two primary levers to address these plateaus: we can either reduce the step size through increased regularisation towards the behaviour policy or decrease the noise by collecting more data per update.

\looseness=-1 Based on this perspective, we show that one simple way to influence PPO's plateauing behaviour---by modulating both of these factors---is to change the number of parallel environments.
However, how best to adjust the other hyperparameters when doing so remains unclear, since more parallel rollouts require either larger minibatches or more optimisation steps, both of which may in turn require adjusting, for instance, the learning rate or regularisation strength~\citep{hilton2022Batch,singla2024sapg}.
We demonstrate that a simple and reliable strategy is to keep the \textit{inner} optimisation process the same: in other words, fix the minibatch size and learning rate, and only increase the number of optimisation steps. 
In a difficult robotics domain, we find that this recipe makes PPO more amenable to massive parallelisation compared to when changing the inner optimisation hyperparameters. 
Finally, we significantly exceed the prior performance ceiling in the challenging 2D physics-based open-ended environment, \texttt{Kinetix}~\citep{matthews2024kinetix}. 
While standard configurations plateau after less than ten billion interactions, scaling PPO to over one million parallel environments allows for sustained monotonic improvement far beyond this point, up to one \textit{trillion} timesteps.

We structure this paper as follows. 
First, \cref{sec:ppo_as_superimposed} empirically justifies our conceptual model of PPO as stochastic optimisation, and shows that both settings share the same mechanisms (e.g., thrashing around a local optimum) and remedies (e.g., reducing the step size) associated with plateaus. We further validate this analogy by showing that changing the outer step size during training is sufficient to either induce a plateau or recover from one.
Next, \cref{sec:what_makes_good_ppo_step} examines the effect of various hyperparameters on the outer step size and update noise and isolates (a) the regularisation strength towards the previous policy, (b) the number of transitions collected per iteration, and (c) the number of optimisation epochs performed on each batch of data as key factors. We then investigate how the optimal step size changes as a function of the training budget, and find that it becomes smaller as the total interaction budget increases. This section ends by arguing that increasing the number of parallel environments is a simple and robust way to lower both the update noise and step size.
\cref{sec:methods} determines how to co-scale the other hyperparameters when increasing parallelisation, and demonstrates the benefits of our recipe in a challenging robotics domain. 
Finally, we showcase significant performance benefits by using our analysis to circumvent learning stagnation in \texttt{Kinetix}.

\section{Background}
\subsection{Proximal Policy Optimisation}
Proximal Policy Optimisation~\citep[PPO]{schulman2017Proximal} is a particularly prevalent, on-policy RL algorithm, with training comprising two distinct phases, which we call the outer and inner loops, respectively.\footnote{These correspond to the data collection and model update steps of standard implementations, see \citet{shengyi2022the37implementation,huang2022cleanrl}.}
In the outer loop, the current policy (also known as the behaviour policy) collects data by rolling out $N_\text{envs}$ parallel environments for $K$ steps each, resulting in a dataset with $N_\text{envs} \cdot K$ transitions.
The inner loop consists of $N_{\text{epochs}}$ passes over this dataset, each pass performing $N_{\text{minibatches}}$ minibatch-SGD gradient steps, usually with the Adam optimiser~\citep{kingma2014adam}.

The agent consists of a policy (also known as the actor) $\pi_\theta(a | s)$, defining a probability distribution over actions for each state, and a critic $V_\theta(s)$, which approximates the sum of discounted returns starting from a particular state $s$ and following $\pi_\theta$. Typically, both of these consist of deep neural networks parametrised by $\theta$, representing either a shared architecture or distinct weights for the actor and critic.
The following loss function is maximised with respect to $\theta$:

\begin{align}
    L_t(\theta) &= \mathbb{E}\left[L_t^\text{CLIP}(\theta) - c_1 L_t^{VF}(\theta) + c_2 \mathcal{H}[\pi_\theta](s_t)\right] \\
    L_t^\text{CLIP}(\theta) &= \min \left( r_t(\theta) \hat A_t, \text{clip}(r_t(\theta), 1 - \epsilon, 1 + \epsilon) \hat A_t \right)  \label{eq:loss_clip} \\
    L_t^{VF}(\theta) &= (V_\theta(s_t) - V_t^\text{target})^2,
\end{align}
where $\mathcal{H}$ is the entropy of the policy, $\hat A_t$ is the advantage calculated using GAE~\citep{schulman2015Highdimensional} and $r_t(\theta) = \frac{\pi_\theta(a_t | s_t)}{\pi_{\theta_\text{behaviour}}(a_t | s_t)}$ is the probability ratio between the current and behaviour policy. $V_t^\text{target}$ is defined as $V_\theta(s_t) + \hat A_t$, computed once before the first optimisation step.
Here, the purpose of the clipping term is to prevent the agent's policy moving too rapidly in any single iteration, and has the effect of zeroing out gradients from transitions where the current policy and the behaviour policy's action probabilities differ by more than $\epsilon$~\citep{schulman2017Proximal}.

\subsection{PPO-EWMA}
In PPO, the behaviour policy serves two different purposes~\citep{schulman2017Proximal,hilton2022Batch}. 
The first is to calculate the \textbf{importance sampling} ratio, in order to correct for the fact that the data-collecting policy is not the same as the current policy being learned (since we do multiple minibatches and epochs per policy update step).
The second is to act as a \textbf{regulariser}, to ensure that the current policy does not drift too far away from a reasonable reference. A key insight from \citet{hilton2022Batch} is that we can use two \textit{different} policies to serve these distinct purposes: the behaviour policy collects a fixed amount of data, and the proximal policy (i.e., the reference policy we regularise towards) is set to the policy from a fixed number of minibatch update steps ago (where older proximal policies lead to stronger regularisation). 
This allows us to control the relative strength of regularisation without altering the data collection process.
Since storing all intermediate policies is expensive in terms of memory, the approximation the authors suggest is an exponentially-weighted moving average (EWMA) of the current policy's weights.

\citet{hilton2022Batch} further argue that the number of parallel environments in standard PPO implicitly influences the regularisation, since it changes the age of the behaviour policy, which in standard PPO is the same as the policy that we regularise towards. 
PPO-EWMA decouples these factors, allowing practitioners to set regularisation independent of parallelisation, and has been used in several recent works to improve training stability, often in asynchronous settings~\citep{hilton2023scaling,zheng2025stabilizing,fu2025areal}.
This is done by modifying the PPO loss function as follows:

\begin{align}
    L_{t,\text{decoupled}}^\text{CLIP}(\theta) &= \mathbb{E}\left[ \frac{\pi_{\mathcolor{red}{\theta_\text{prox}}}(a_t | s_t)}{\pi_{\theta_\text{behaviour}}(a_t | s_t)} \min \left( \mathcolor{red}{r^\text{prox}_t(\theta)} \hat A_t, \text{clip}(\mathcolor{red}{r^\text{prox}_t(\theta)}, 1 - \epsilon, 1 + \epsilon) \hat A_t \right) \right]  \label{eq:loss_clip_ewma},
\end{align}

where $\mathcolor{red}{r^\text{prox}_t(\theta)} = \frac{\pi_\theta(a_t | s_t)}{\pi_{\mathcolor{red}{\theta_\text{prox}}}(a_t | s_t)}$ and $\theta_\text{prox}$ is an EWMA of $\theta$, updated after every minibatch as $\theta_\text{prox} \leftarrow \beta_\text{prox} \theta_\text{prox} + (1 - \beta_\text{prox}) \theta$.
The first ratio now is for importance sampling whereas $\mathcolor{red}{r^\text{prox}_t(\theta)}$ controls regularisation. Importantly, if $\epsilon=\infty$ and no clipping happens, the product of the ratios $\frac{\pi_{\mathcolor{red}{\theta_\text{prox}}}(a_t | s_t)}{\pi_{\theta_\text{behaviour}}(a_t | s_t)} \cdot \frac{\pi_\theta(a_t | s_t)}{\pi_{\mathcolor{red}{\theta_\text{prox}}}(a_t | s_t)}$ recovers the standard importance sampling ratio $r_t(\theta)$~\citep{hilton2022Batch}.

For the EWMA, the center of mass (COM) is defined as $\frac{1}{1 - \beta_\text{prox}} - 1$, measured in minibatch update steps. This quantity controls the ``age'' of the reference policy, and thereby the regularisation strength. 
As an example, $\beta_\text{prox}=\frac{1024}{1025} \approx 0.99902$ corresponds to a center of mass of $1024$, meaning that the average age of each term in the EWMA calculation is $1024$ minibatches. In other words, the regularisation here is comparable to normal PPO when using $256$ minibatches and $8$ epochs: because standard PPO refreshes the behaviour policy every $2048$ minibatches ($8 \times 256$), the average age of its regularisation target is likewise $1024$ minibatches.

For some experiments in this paper, we use PPO-EWMA as an analysis tool which provides a more interpretable and granular way to control the regularisation strength compared to directly altering the clipping threshold. We also find that low COMs lead to more stable training than high $\epsilon$'s, allowing us to more easily study the effects of weak regularisation.
However, \cref{app:beta_vs_epsilon} shows that we can largely counteract changes to the COM by appropriately adjusting $\epsilon$, and vice versa, confirming that these hyperparameters affect the same underlying mechanism. Furthermore, our final results in \cref{sec:methods} use standard PPO, showing that our insights transfer to the more commonly-used variant.

\section{PPO as a Stochastic Optimisation Process}\label{sec:ppo_as_superimposed}
In this section we empirically justify our conceptual model of PPO's outer loop as stochastic optimisation. For these experiments, we use a state-based robotic locomotion task comprising $512$ procedurally-generated morphologies built using the \texttt{Jax2D} physics engine~\citep{matthews2024kinetix} and a simple noisy convex optimisation problem: minimizing $\textbf{x}^T\textbf{x}$, with $\textbf{x} \in \mathbb{R}^{50}$. We perform standard gradient descent, but add Gaussian noise with standard deviation $\frac{3}{\sqrt{50}}$ to the gradients. To more clearly demonstrate the similarities to RL, we plot the negative of the euclidean distance to the optimal solution $\textbf{x}^* = \textbf{0}$.
For all figures, we plot the mean and shade the 95\% CI over 5 seeds  unless otherwise noted. See \cref{app:env_details} for more details and hyperparameters.

\begin{figure}[h]
    \centering
    \begin{subfigure}[t]{0.45\linewidth}
        \centering
        \includegraphics[width=1\linewidth]{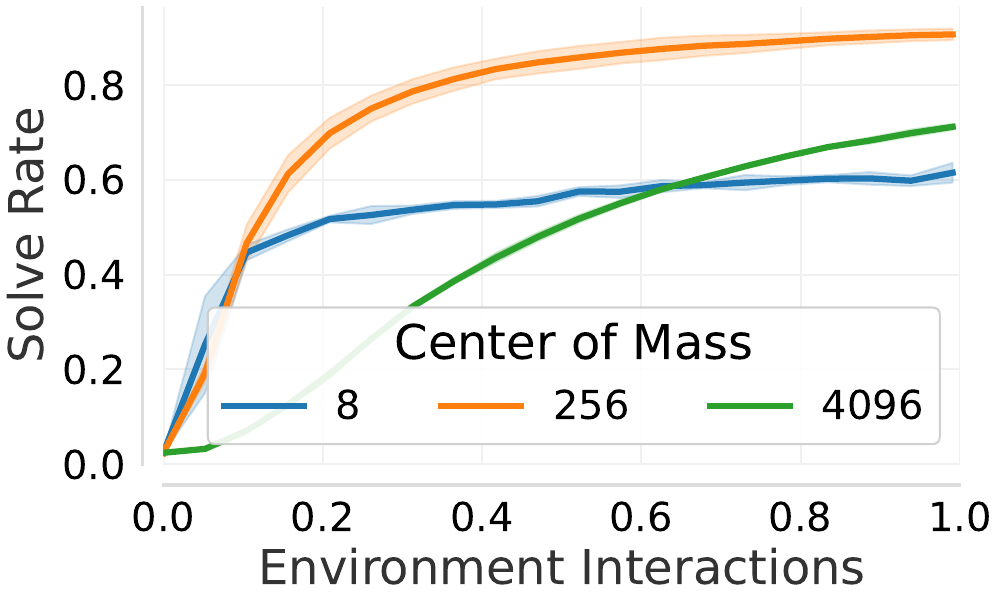}
        \caption{PPO}
        \label{fig:thrash_solverate}
    \end{subfigure}\hfill%
    \begin{subfigure}[t]{0.45\linewidth}
        \centering
        \includegraphics[width=1\linewidth]{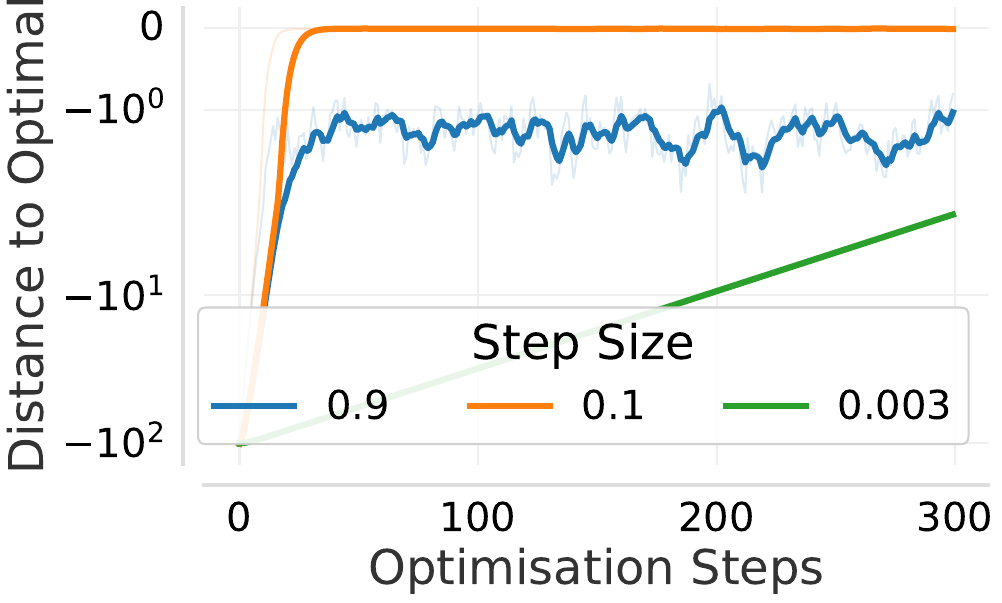}
        \caption{Noisy Convex Optimisation}
        \label{fig:thrash_solverate_sgd}
    \end{subfigure}
    \caption{Comparing the behaviour in (a) PPO and (b) convex optimisation with stochastic gradients. In (a) having too large of an outer step size (in particular, having a center of mass of the proximal policy being too low) leads to a suboptimal plateau, with the same behaviour occurring in (b). Solve rate corresponds to the policy's average success rate over all 512 morphologies.}
    \label{fig:thrash_solve_both}
\end{figure}

We start with the observation that initially inspired our conceptual model: scaling the outer-loop step size (regularisation strength in this case) in PPO has a similar effect on the resulting learning curves as scaling the learning rate in SGD.
When it is too high, e.g., the blue lines in \cref{fig:thrash_solve_both}, performance plateaus at a suboptimal level, and when it is too low (the green lines), the optimisation process fails to converge within the allocated budget.

\subsection{Learning Dynamics Under Excessive Step Size}
\begin{figure}[h]
    \begin{subfigure}[t]{0.33\linewidth}
        \centering
        \includegraphics[width=1\linewidth]{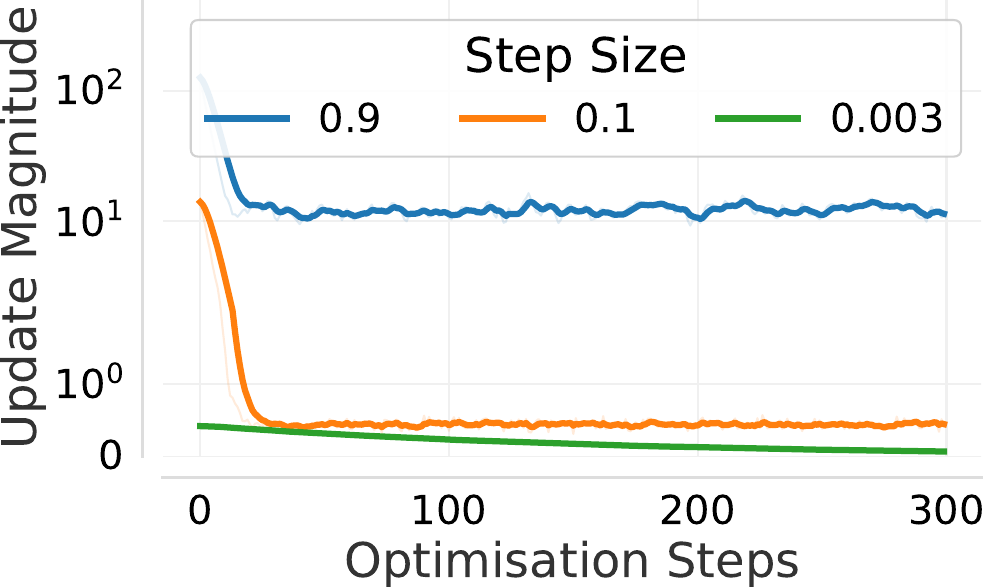}
        \caption{Update size (SGD)}
        \label{fig:thrash_sgd}
    \end{subfigure}\hfill%
    \begin{subfigure}[t]{0.33\linewidth}
        \centering
        \includegraphics[width=1\linewidth]{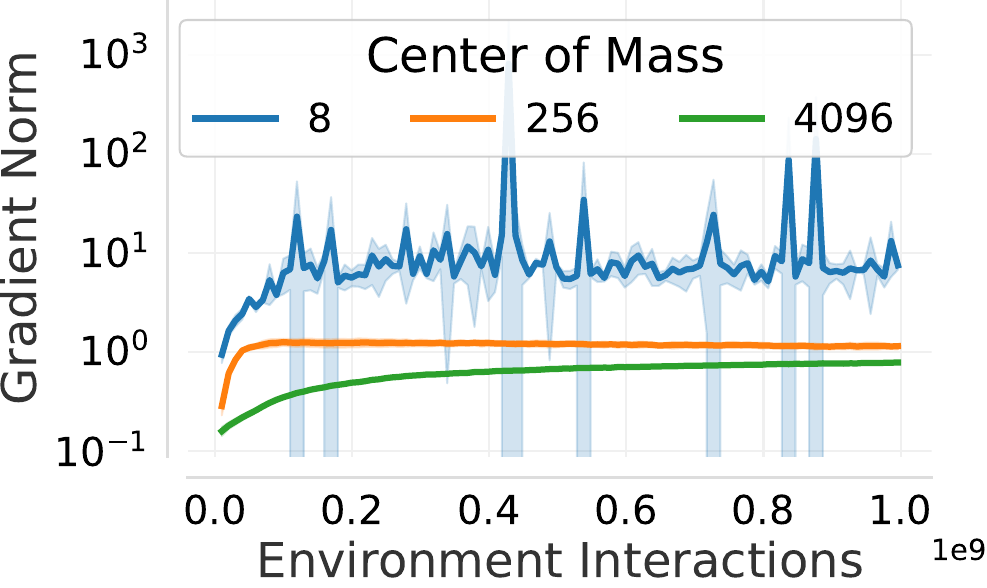}
        \caption{Gradient Norm}
        \label{fig:thrash_gradnorm}
    \end{subfigure}\hfill%
    \begin{subfigure}[t]{0.33\linewidth}
        \centering
        \includegraphics[width=1\linewidth]{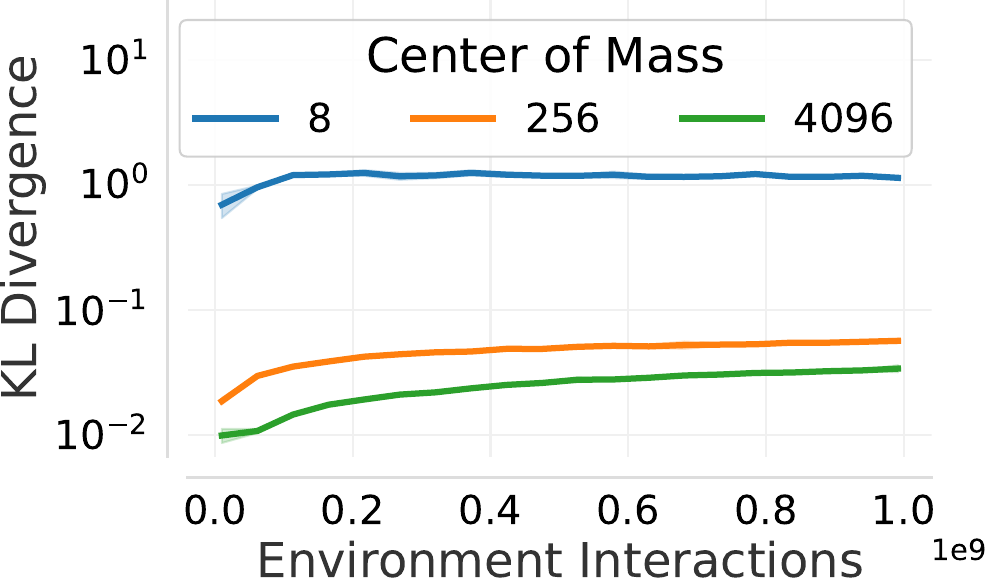}
        \caption{KL (Behavior Policy)}
        \label{fig:thrash_kl_behav}
    \end{subfigure}
    \caption{
    (a) In stochastic optimisation, the update magnitude is consistently large when the step size is too large, despite a stagnating loss.
    (b,c) Showing that PPO shares similar dynamics.
    }
    \label{fig:thrashing}
\end{figure}

While the similarity in the learning curves is striking, it does not provide insight into the mechanisms by which the outer loop step size influences learning progress. As illustrated in \cref{fig:thrash_sgd}, large learning rates in gradient descent induce updates that bounce around the local minimum, experiencing large gradient norms but no decrease in the loss.
\cref{fig:thrash_gradnorm,fig:thrash_kl_behav} show an analogous effect in PPO agents: large outer-loop step sizes due to weak regularisation result in performance stagnating despite large policy updates and gradient norms.\footnote{While the raw gradient norm is high, we perform standard gradient clipping whenever the norm is above $0.5$.}  This suggests that the performance plateau is caused by thrashing around a local optimum rather than converging to a suboptimal stationary point.

\begin{figure}[h]
    \centering
    \begin{subfigure}[t]{0.45\linewidth}    
        \includegraphics[width=1\linewidth]{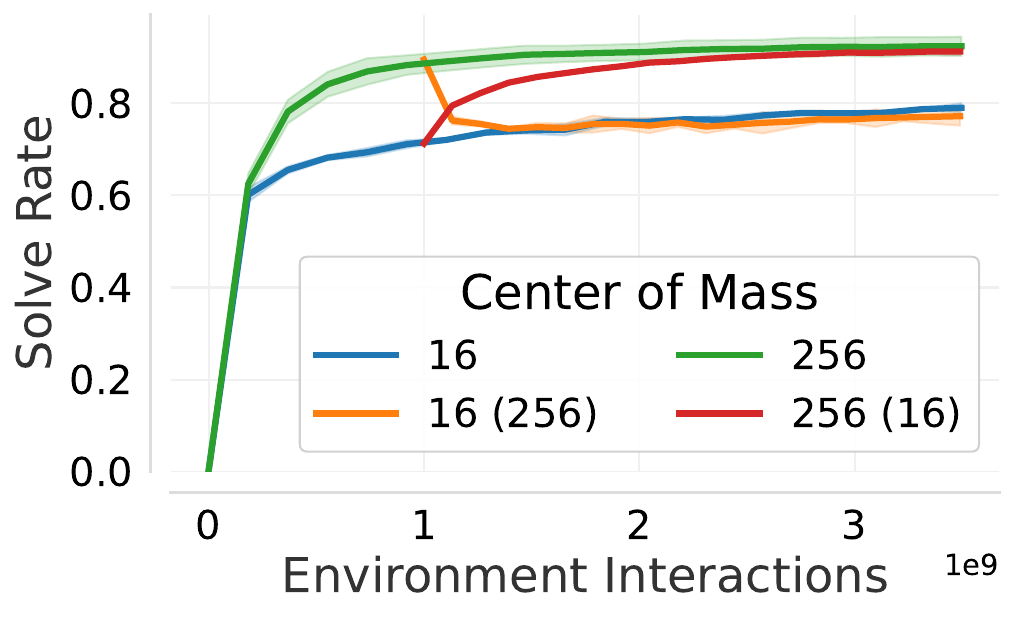}
        \caption{PPO-EWMA}
        \label{fig:not_plasticity}
    \end{subfigure}\hfill
    \begin{subfigure}[t]{0.45\linewidth}    

        \includegraphics[width=1\linewidth]{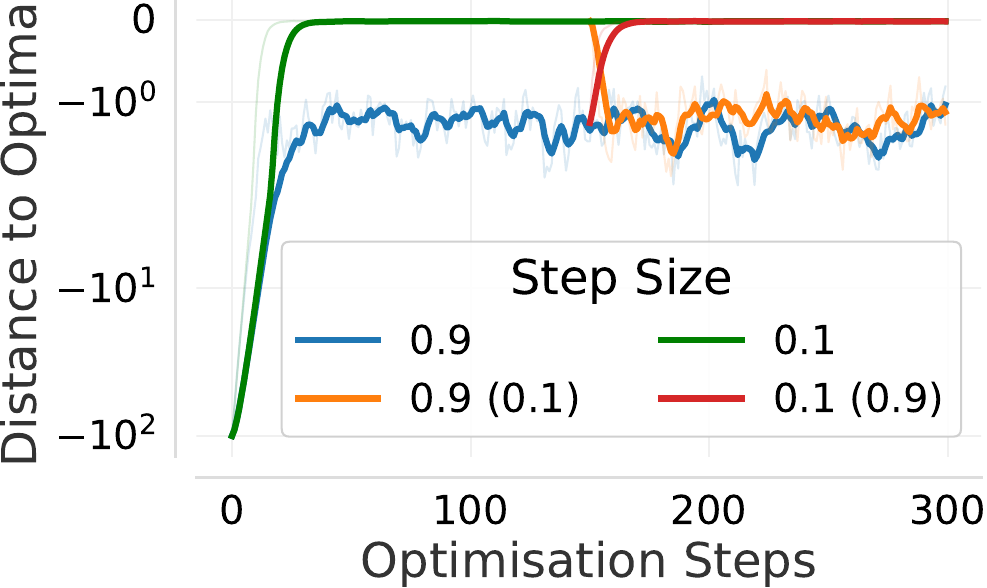}
        \caption{SGD}
        \label{fig:not_plasticity_sgd}
    \end{subfigure}
    \caption{(a) Loading checkpoints and retraining with a different COM recovers the performance of the most recent regularisation strength. The legend indicates the center of mass, and the number in brackets indicates the starting COM. (b) The same phenomenon occurs in stochastic optimisation.}\label{fig:not_plasticity_both}
\end{figure}

We next confirm that these plateaus are a direct consequence of the outer step size rather than the policy network being unable to learn or the generated data being insufficient to learn from. 
\cref{fig:not_plasticity} shows that increasing the proximal policy's COM (thereby reducing the outer step size) after the agent has plateaued allows it to immediately resume learning, ultimately recovering the same asymptotic performance as the higher COM. Moreover, if we reduce the COM, performance drops to the suboptimal plateau associated with the larger step size, exactly matching the behaviour of increasing the learning rate in noisy stochastic optimisation, shown in \cref{fig:not_plasticity_sgd}.

\subsection{Decoupling the Inner and Outer Loops}\label{sec:both_loops_matter}
Having established the importance of the outer step size in influencing an agent's plateauing behaviour, we close off this section by comparing how different properties of the inner and outer loop differ, and which hyperparameters influence each process in \cref{tab:ppo-bilevel}.

\begin{table}[h]

    \centering
    \caption{The differences between properties of the inner loop (parameter-space updates) and the outer loop (policy-space updates) in PPO. $J(\pi)$ is defined as the expected discounted return of $\pi$.}
    \label{tab:ppo-bilevel}
    \begin{tabular}{lll}
        \toprule
        Property      & Inner loop                   & Outer loop                       \\
        \midrule
        Learning rate & Adam learning rate $\eta$       & Regularisation (COM, $\epsilon$) \& epochs   \\
        Noise         & Minibatch vs rollout batch           & Rollout batch vs true gradient\\
        Curvature     & Neural network hessian $\|H_\theta\|$ & Policy landscape $\|\nabla_{\pi} J(\pi)\|$        \\
        \bottomrule
    \end{tabular}
\end{table}

\begin{figure}[h]
    \centering
    \includegraphics[width=1\linewidth]{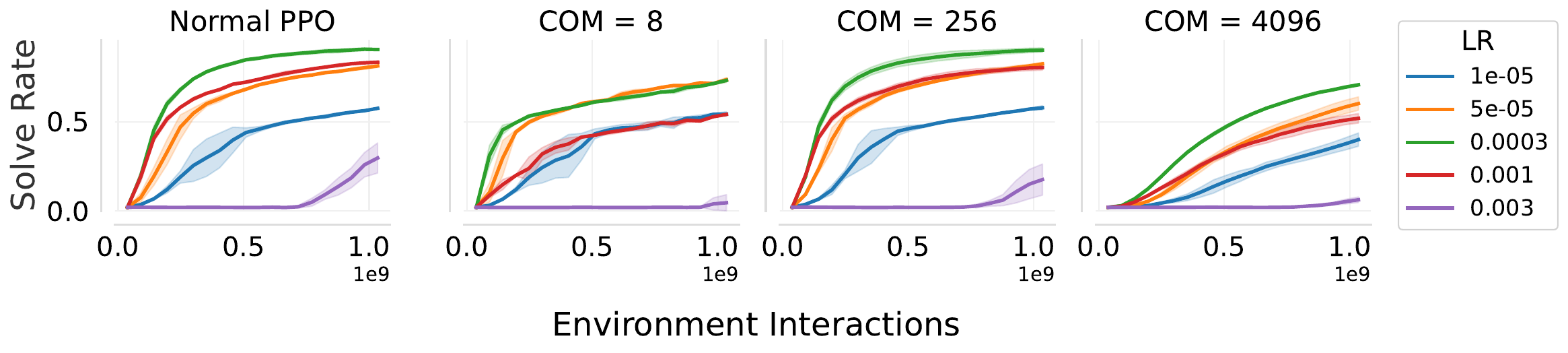}
    \caption{Tuning the learning rate cannot counteract a poor outer step size. Here we sweep over whether or not to anneal LR for each run, and show the best result per learning rate.}
    \label{fig:lr_cannot_counteract_beta}
\end{figure}

Furthermore, to demonstrate that changing the inner loop cannot always compensate for a poor outer loop step size, we tune the learning rate (and sweep over annealing vs not annealing it to zero over the course of training) separately for each center of mass and show the results in \cref{fig:lr_cannot_counteract_beta}. 
Here we can see that if we have an outer loop step size that is too large (with COM = 8), then regardless of the learning rate, performance is still significantly worse than when we tune the COM appropriately. 
Further, we see that the same learning rate is roughly optimal for all COMs, showing that these two hyperparameters affect different learning mechanisms and, importantly, are not interchangeable.

\FloatBarrier
\section{Understanding PPO's Outer Loop}\label{sec:what_makes_good_ppo_step}
Having established PPO's similarities to stochastic optimisation, we next focus on understanding which hyperparameters modulate the outer step size and noise level.
We first show in \cref{sec:modulates_reg} that \textit{regularisation} towards previous policies directly influences the outer step size; next, in \cref{sec:modulates_epochs}, we demonstrate that the number of optimisation epochs we perform also controls the agent's plateauing behaviour; in \cref{sec:modulates_batch_size}, we show that, similarly to SGD, the update noise matters too, and larger batch sizes admit higher step sizes without plateauing, whereas smaller batch sizes are very susceptible to overly large steps. Finally, \cref{sec:how_choose_step_size} suggests some rules of thumb for how to set the step size appropriately, and how this depends on the available computational budget.

\subsection{Regularisation}\label{sec:modulates_reg}
\begin{figure}[h]
    \begin{subfigure}[t]{0.5\linewidth}
        \centering
        \includegraphics[width=1\linewidth]{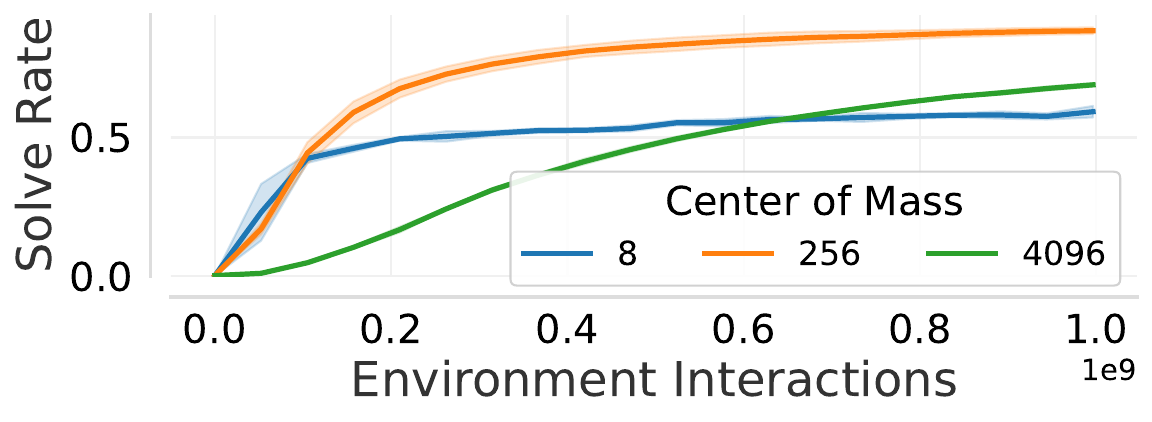}
        \caption{Changing COM}
        \label{fig:com_causes_plateaus}
    \end{subfigure}%
    \begin{subfigure}[t]{0.5\linewidth}
        \centering
        \includegraphics[width=1\linewidth]{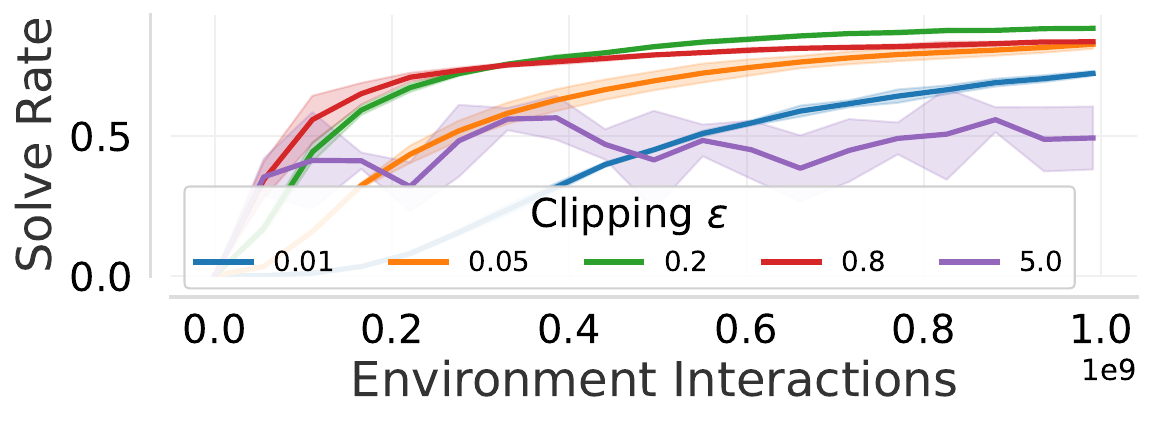}
        \caption{Changing $\epsilon$}
        \label{fig:epsilon_causes_plateaus}
    \end{subfigure}%
    \caption{Weak regularisation, corresponding to either (a) too low of a COM or (b) too large of a clipping $\epsilon$ can lead to premature plateaus.}\label{fig:epsilons_causes_plateaus_both}
\end{figure}

We consider two ways to control regularisation: either altering the center of mass of the proximal policy in PPO-EWMA (controlling how old the policy is we regularise towards) or changing PPO's clipping $\epsilon$ parameter. 
As shown in \cref{fig:epsilons_causes_plateaus_both}, weak regularisation (either via a high $\epsilon$ or low COM) leads to premature plateaus, whereas overly strong regularisation (low $\epsilon$ or high COM) leads to slow learning relative to the available environment sample budget.

\FloatBarrier
\subsection{Optimisation Epochs}\label{sec:modulates_epochs}
\begin{figure}[h]
    \centering
    \includegraphics[width=1\linewidth]{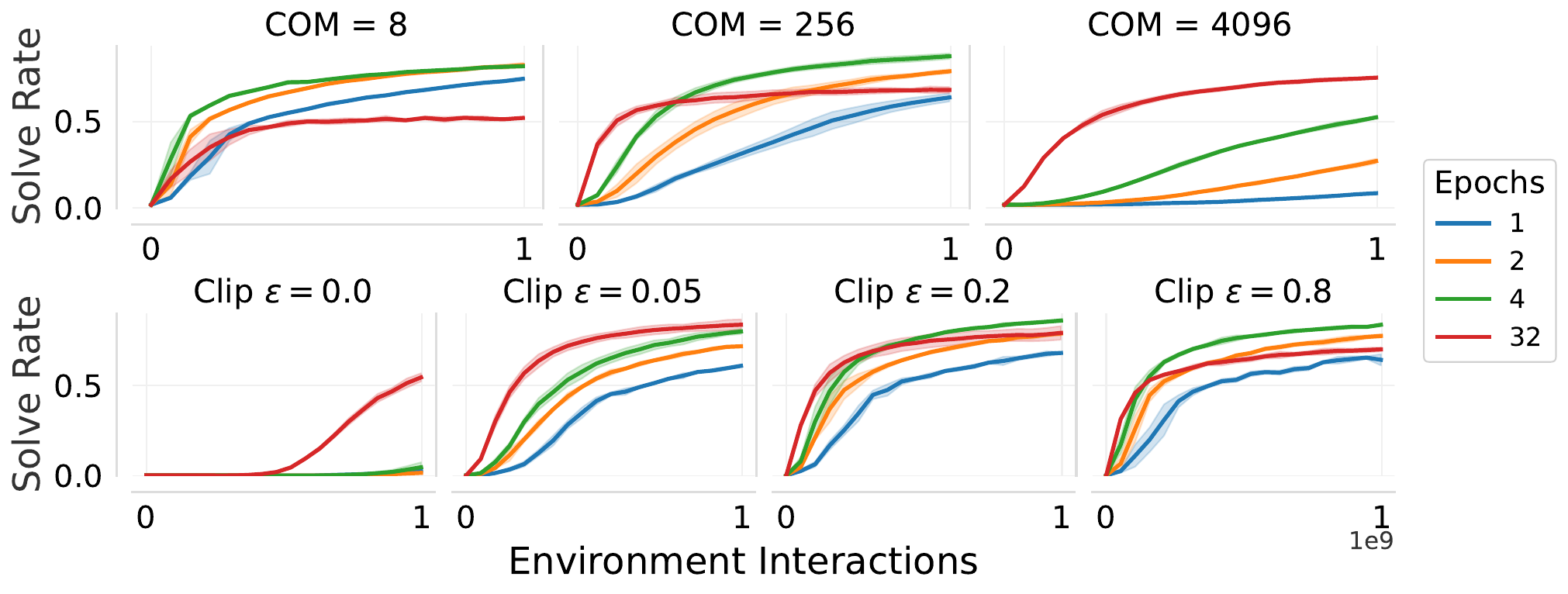}
    \caption{How changing number of epochs influences (top) PPO-EWMA and (bottom) normal PPO. Stronger regularisation can partially alleviate plateaus due to too many epochs, thereby reaching higher asymptotic performance.
    See \cref{app:env_details} for full hyperparameters and experimental details.
    }
    \label{fig:epochs_counteract_beta_and_eps}
\end{figure}

In \cref{fig:epochs_counteract_beta_and_eps}, we show that changing the number of inner optimisation epochs we perform on each batch of data also has a direct effect on an agent's plateauing behaviour. This is again dependent on the strength of the regularisation, where weak regularisation plateaus more easily when increasing the number of epochs, and strong regularisation learns slowly with a small number of epochs.
One notable result in bottom left panel is that even with a clipping term of $\epsilon = 0$, a large number of epochs can be beneficial. This in particular is likely due to the Adam momentum term, and the fact that the PPO update can overshoot the $\epsilon$ threshold~\citep{ilyas2018deep,pmlr-v115-wang20b}. Specifically, clipping only zeros out the gradients once the ratio \textit{already} exceeds the threshold, so if the first update step is large (either because of a large (inner) learning rate, or a large momentum term as is the case here), the ratio after this initial update can be far outside the $1 \pm \epsilon$ range. See \cref{app:beta_vs_epsilon} for more details.

\FloatBarrier
\subsection{Rollout Batch Size}\label{sec:modulates_batch_size}
Following from the analogy to stochastic optimisation, if the core problem is that we take steps that are too large on noisy targets, then another solution would be to increase the signal-to-noise ratio via larger batches~\citep{smithDont,mccandlish2018empirical}. To investigate this, we compare the performance of agents when varying the number of parallel environments, and keeping the same number of minibatches---meaning we change only the minibatch size, keeping everything else constant.
Further, to isolate the impact of update \textit{quality}, we compare agents based on the number of policy update steps;  therefore, agents with larger batches do see more data, and the comparison is not fair in terms of environment transitions. Nevertheless, it allows us to analyse how much the quantity of data per update step changes the effect of regularisation. 

\begin{figure}[h]
    \centering
    \includegraphics[width=1\linewidth]{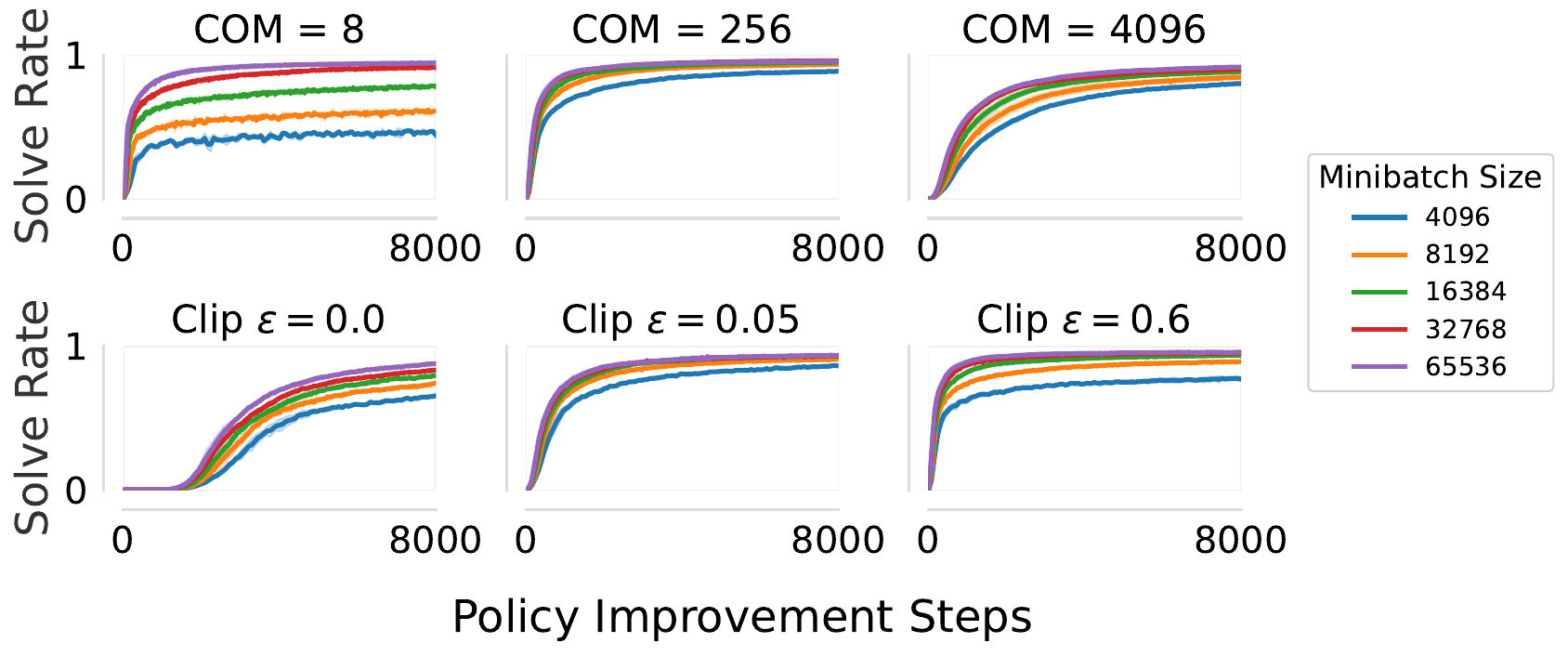}
    \caption{Showing the effect of larger minibatches when changing the (top) COM of $\theta_\text{prox}$ in PPO-EWMA; and (bottom) clipping $\epsilon$ term in standard PPO. Here the x-axis is the number of policy update steps. Larger batches are less susceptible to plateauing when paired with weak regularisation.}
    \label{fig:larger_bs_same_lr_both}
\end{figure}

\cref{fig:larger_bs_same_lr_both} shows that larger batch sizes are significantly more robust to weaker regularisation than smaller ones.
For instance, if we have a small batch size (e.g., $4096$), weak regularisation (e.g. COM of 8 or $\epsilon = 0.6$) performs significantly worse compared to when it is paired with a larger batch size.\footnote{While these minibatches may seem large, they are consistent with recent work on hardware-accelerated environments, which achieve speedups via parallelisation~\citep{makoviychuk2021Isaac,nikulin2023xlandminigrid}}
This suggests that the higher signal-to-noise ratio achieved through larger batches admits larger outer step sizes without plateauing, again echoing known results from stochastic gradient descent~\citep{krizhevsky2014one,goyal2017accurate,smithDont,mccandlish2018empirical}.

\FloatBarrier
\subsection{Choosing an Appropriate Step Size}\label{sec:how_choose_step_size}
\begin{figure}[H]
    \centering
    \includegraphics[width=1\linewidth]{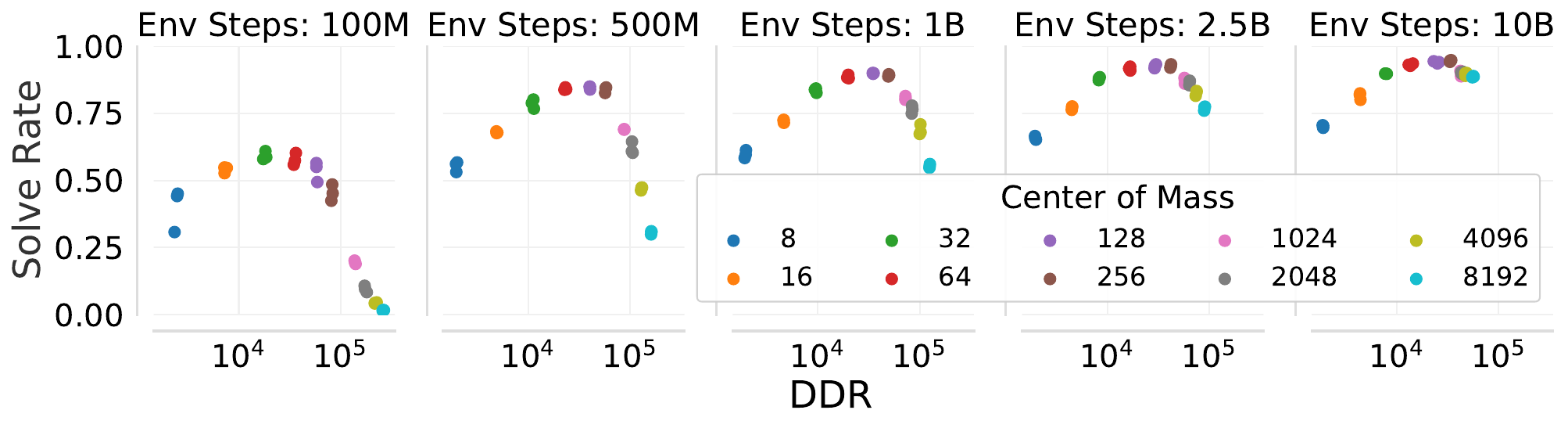}
    \caption{DDR (averaged over training) vs. the maximum solve rate achieved in that run for various compute budgets. Each dot is a single training run and different dots of the same colour are different random seeds. The dashed red line indicates the approximate performance ceiling.}
    \label{fig:ddr_compute}
\end{figure}

The analysis from this section suggests that the important factors influencing whether or not an agent plateaus at a suboptimal performance ceiling are (a) the number of transitions we use per policy update step, and (b) the size of the deviation from the reference policy. We next turn to the question of how to set the corresponding hyperparameters to reasonable values.
To do so, we unify (a) and (b) into the Data to Divergence Ratio (DDR): the number of data points per unit KL divergence from the behaviour policy. 
\cref{fig:ddr_compute} shows that performance suffers at both ends of the DDR spectrum; however, the mechanisms underlying this behaviour are distinct for each extreme.
Low DDR values lead to early plateaus (and therefore do not improve much when given additional training time), whereas high DDR values cause learning to progress slowly, meaning that these agents fail to reach their performance ceiling within the fixed sample budget. However, since slower learning is acceptable under larger interaction budgets, our results suggest that \textit{as we increase the training budget, we should increase the DDR accordingly to avoid a premature plateau.}

Having established the importance of increasing the DDR as we train for more samples, we propose that one simple way to do so is to increase the number of parallel environments. This directly increases the amount of data per policy improvement step (thereby reducing the update noise) and indirectly lowers the outer step size due to the behaviour policy age, measured in environment samples, increasing~\citep{hilton2022Batch}.
However, it remains unclear how we should adjust the other hyperparameters when increasing parallelisation, and we address this question in the next section.

\section{A Reliable Recipe for Scaling Parallelisation in PPO}\label{sec:methods}
Our results thus far suggest that we need to scale down the outer step size as our computational budget increases in order to avoid premature stagnation. In addition, increasing the number of parallel environments is a desirable way to do so, since it reduces both the step size and the update noise, while allowing more samples to be processed within the same amount of wall-clock time. 
However, when we increase the number of parallel environments in PPO, we have more data per policy update step, and this necessitates adjusting some of the other hyperparameters. There are three primary ways to partition this data:
\begin{enumerate}
    \item Have more minibatches of the same size.
    \item Have larger minibatches with the same learning rate.
    \item Have larger minibatches, and scale the learning rate according to the square-root rule for Adam~\citep{krizhevsky2014one,malladi2022sdes,granziol2022learning,hilton2022Batch}.
\end{enumerate}

\begin{figure}[h]
    \centering
    \includegraphics[width=1\linewidth]{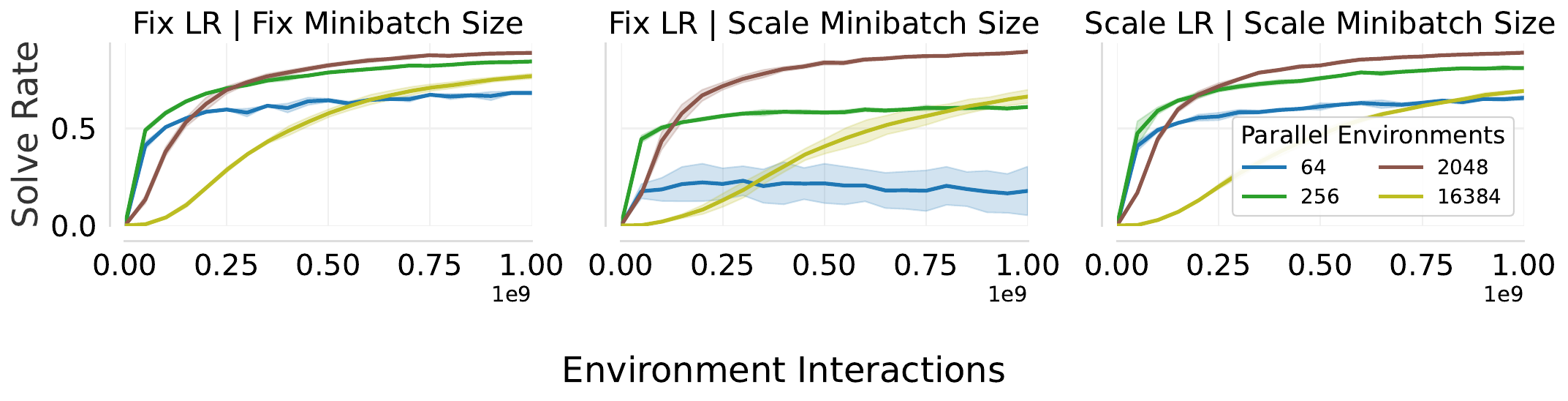}
    \caption{Comparing different approaches when varying $N_\text{envs}$. Keeping the inner optimisation process unchanged performs best, whereas performance of small minibatches suffers when scaling the minibatch size without adjusting the learning rate. See \cref{app:additional_scaling_results} for more granular plots.}
    \label{fig:how_to_scale_ppo}
\end{figure}

\cref{fig:how_to_scale_ppo} shows that having more minibatches while keeping everything else fixed works reliably. 
This recipe preserves the dynamics of the inner optimisation process---with an unchanged learning rate, minibatch size, etc., and merely changes the number of optimisation steps we do. This strategy is further justified by \cref{sec:ppo_as_superimposed}, where we show that the optimal inner and outer step sizes are largely independent. 
However, larger minibatches (with a scaled learning rate) tend to result in better hardware utilization, and thus faster training.\footnote{This is typically beneficial when compute bound (e.g. by using larger models), whereas in our sample-bound locomotion task, the wall-clock gains are marginal (see \cref{fig:ppo_scaling_sps} in \cref{app:additional_scaling_results}).} 
While larger minibatches can work well in certain environments, \cref{fig:sapg_scale_results} shows that they sometimes lead to training instability and lower plateaus~\citep{do2024revisiting,su2025characterization}.
In summary, we recommend a stability-first procedure: increase the number of minibatches while keeping the learning rate and minibatch size fixed. Only increase minibatch size (adjusting the learning rate appropriately) if hardware utilization is a bottleneck.

\subsection{Robotics Results}\label{sec:isaacgymresults}
\begin{figure}[H]
    \centering
    \includegraphics[width=1\linewidth]{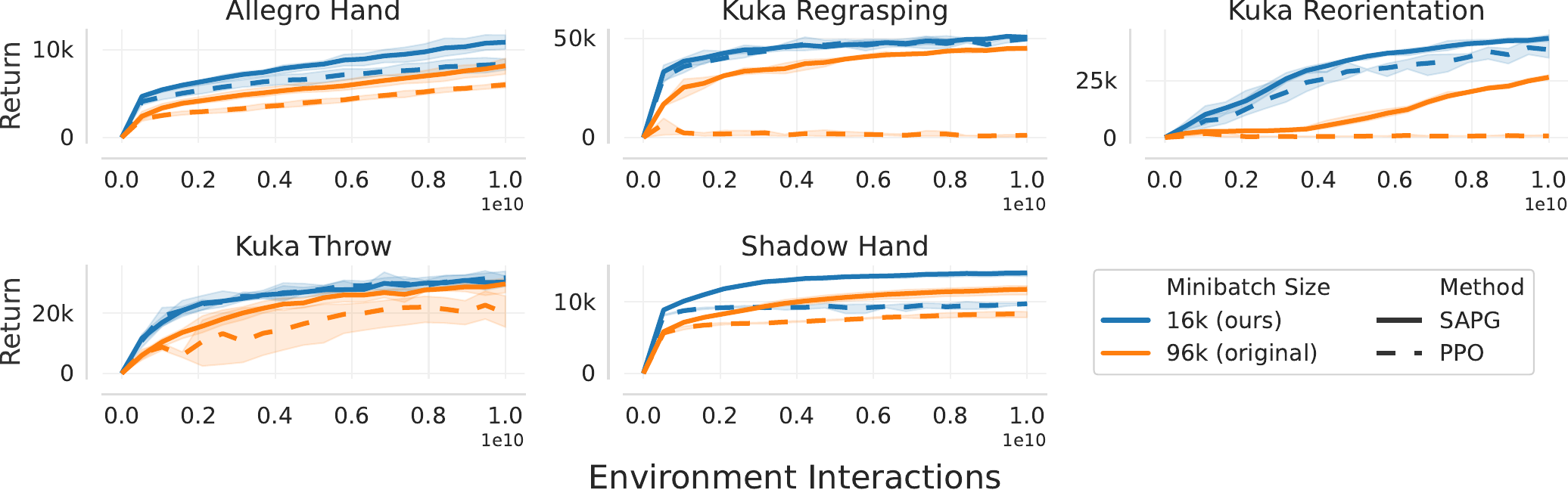}%
    \caption{We take the code from \citet{singla2024sapg}, and make \textbf{one} change---setting the minibatch size to 16k (which is the default in Isaacgym) instead of using 96k like \citet{singla2024sapg} do. 
    When using our recommendations, vanilla PPO performs much better across the board, and the gap between it and SAPG is reduced. 
    Furthermore, SAPG also benefits from the same change.}
    \label{fig:sapg_scale_results}
\end{figure}

To demonstrate the practical utility of our scaling recipe, we consider a set of difficult robotics tasks from Isaacgym~\citep{makoviychuk2021Isaac} used by \citet{singla2024sapg}. 
The default minibatch size for several tasks in Isaacgym is $16384$.\footnote{\url{https://github.com/isaac-sim/IsaacGymEnvs/blob/main/isaacgymenvs/cfg/train/AllegroHandLSTM_BigPPO.yaml\#L88}}
However, when increasing parallelisation, \citet{singla2024sapg} fix the learning rate and increase the minibatch size to $4
\times$ the number of parallel environments---resulting in a minibatch size of $98304$ for $24576$ environments. 
Following our recommendations, we make a single adjustment: we revert the minibatch size back to the default $16384$ for both PPO and SAPG, the new method introduced by \citet{singla2024sapg}.
\cref{fig:sapg_scale_results} shows that this one change significantly outperforms the default setting for both methods, makes PPO more amenable to additional parallelisation, and reduces the performance gap between it and SAPG.

\section{Batch Size Scaling Enables Open-Ended Learning}\label{sec:kinetix_results}
Finally, we show that, by using our analysis, we can overcome learning stagnation in the challenging open-ended domain of \texttt{Kinetix}~\citep{matthews2024kinetix}. In this setting, agents train on a procedurally-generated distribution of tasks with the objective of achieving robust generalisation over the entire space of tasks. 
There are three different training distributions (\texttt{\textbf{s}mall}, \texttt{\textbf{m}edium} or \texttt{\textbf{l}arge}), corresponding to the maximum number of entities there are in the scene; each of these is treated as a separate experiment.
The best performing approach on \texttt{Kinetix} is SFL~\citep{rutherford2024noRegrets}---an autocurriculum method that samples training tasks that have high learnability (i.e., those where the agent has about a 50\% chance of success, meaning they are neither too easy nor too hard).
Like much of the field of Unsupervised Environment Design, SFL uses PPO as the underlying learning algorithm~\citep{dennis2020Emergent,jiang2021prioritized,holder2022Evolving}.
\begin{figure}[h]
    \centering
    \includegraphics[width=1\linewidth]{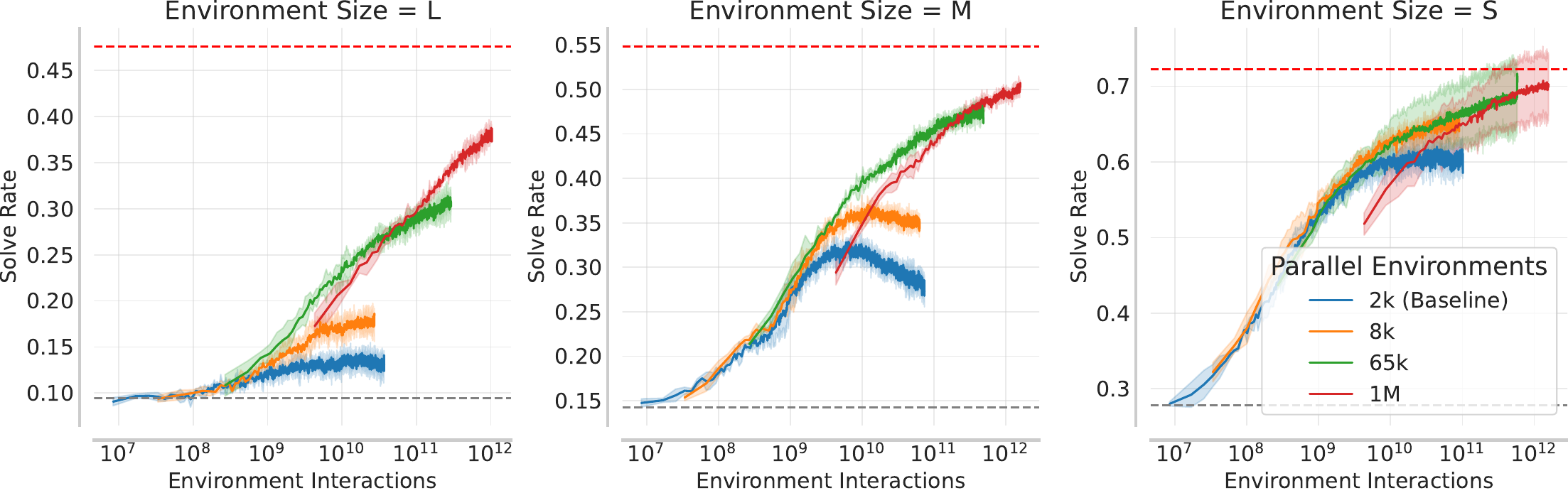}
    \caption{SFL on \texttt{Kinetix}, showing that increasing the number of parallel environments maintains performance improvement for much longer. The dashed red line is an approximation of optimal performance, since not all sampled environments are solvable, while the grey line indicates a random policy's performance. We plot mean and 95\% CI over 3 seeds. The curves are truncated at different x-values since using fewer parallel environments takes a much longer wall-clock time to generate a particular number of transitions. We run the baseline from \citet{matthews2024kinetix} for longer to clearly show the performance degradation.}
    \label{fig:sfl_main_results}
\end{figure}

As a way to measure how well the agent is performing on the full distribution of tasks, we calculate performance on a fixed set of environments randomly sampled from the training distribution.
In \cref{fig:sfl_main_results} we show that the default configuration used by \citet{matthews2024kinetix} plateaus early, and its performance even starts degrading when given more samples. This means that any additional compute is effectively wasted. 
However, by simply increasing the number of parallel environments (using our scaling recipe),\footnote{For wall-clock reasons, at the cost of some learning efficiency, we use a hybrid approach, where we have $32\times$ the number of minibatches, each minibatch is $16\times$ the size, and the learning rate is $4\times$ larger. See \cref{app:sfl_scaling} for more details.} we are able to sustain performance improvement for much longer, ultimately reaching significantly higher performance. 
This effect is more pronounced in the more difficult and wider \texttt{large} distribution of tasks, whereas 65k environments seems sufficient for the \texttt{small} setting.
Importantly, we find that we can reliably scale to over 1M parallel environments ($512\times$ more than \citet{matthews2024kinetix} used) across 128 GPUs, and this level of parallelisation is imperative in order to be able to collect more than a trillion environment transitions within a reasonable wall-clock time. As predicted by the shifting optima in \cref{fig:ddr_compute}, such large training budgets require the small, high-quality updates provided by scaling up to avoid premature plateaus.

We note that as we use additional GPUs, we process more environments in the filtering stage (since this is effectively free in terms of wall-clock time); however, \cref{app:sfl_ablations} shows that this alone is insufficient to prevent stagnation unless paired with increased training parallelisation.

\FloatBarrier
\section{Related Work}

There has been a long line of work that demonstrates scaling RL tends to be much more difficult and less straightforward than supervised learning~\citep{ota2021training,bjorck2021towards,schwarzer2023bigger,ceron2024mixtures,lee2025simba,lee2025hyperspherical,rybkin2025valuebased,wang20251000}.
While model scaling is a common topic of study~\citep{lee2025simba,lee2025hyperspherical}, less focus has been put on data scaling, i.e., what happens when we give agents a massive amount of \textit{online} experience. 
Part of this has been due to the difficulty in running these experiments; however, the recent wave of GPU-accelerated RL environments has made it feasible to study these questions with only modest hardware requirements~\citep{brax2021Github,nikulin2024Xland100B,bonnet2024jumanji,matthews2024Craftax,matthews2024kinetix}.
On this topic, \citet{bharthulwar2025staggered} demonstrate that one benefit of parallelisation is an increase in data diversity, but that this is not always the case when all parallel environments are very similar. 
They propose to stagger their resets, so that each worker simulates temporally unrelated chunks of experience. This provides an added explanation for why increasing the parallelisation is so effective in \texttt{Kinetix}---since nearly every parallel environment is simulating a unique environment, the diversity of experience is directly influenced by the parallelisation. Relatedly, \citet{mclean2025multitask} show that increasing the number of tasks an agent trains on can reduce plasticity loss, but they consider only the 10 and 50 tasks from MetaWorld~\citep{yu2020meta}, significantly less than the millions of tasks we train on.
Finally, \citet{mayor2025Impact} investigate the tradeoffs between scaling parallelisation and rollout length in PPO, and find that for a fixed data budget, more parallelisation tends to be preferred. By contrast, we find that \textit{more} data per rollout batch has benefits.

Our work shows that some of the instabilities of PPO are similar to pathologies from classic optimisation theory~\citep{robbins1951stochastic,luenberger1984linear,bertsekas1997nonlinear}, and therefore many remedies from that field are related~\citep{polyak1964some,nocedal2006numerical}.
However, we are not the first to view PPO's outer step as analogous to an optimisation problem. \citet{tan2024beyond} directly view the difference in parameters across successive PPO updates as a ``gradient'', and either use momentum or apply the update with an outer ``learning rate'' that is not equal to $1$. By contrast, we use our conceptual model to understand PPO's behaviour better, thereby finding better ways to train agents that do not stagnate without altering the underlying, tried and tested, algorithm.

This paper is also related to two-timescale analyses of reinforcement learning, a field that includes learning nonlinear representations and linear value functions at different timescales~\citep{chung2018two}, and the decoupled optimisation of the actor and critic~\citep{wu2020finite,zeng2024two,zeng2024fast}.
However, instead of analysing theoretical implications or developing new algorithms, we investigate the empirical similarities between PPO and stochastic optimisation, and how this provides practical guidance for preventing premature plateaus.

Another related field is that of open-endedness, where the goal is to obtain an algorithm that we can run forever, and that will continually result in novel artifacts~\citep{stanley2019open,dennis2020Emergent,team2021Openended,holder2022Evolving,team2023Humantimescale}.
However, one prerequisite for such an algorithm is agents that do not stagnate~\citep{hughes2024Openendedness}---which we have shown can happen in several different domains, and that increasing parallelisation is one way to alleviate this issue.

Finally, our work sheds some light on the folk knowledge that higher parallelisation levels tend to result in faster wall-clock training times, at the cost of worse sample efficiency~\citep{brax2021Github,makoviychuk2021Isaac}. As \citet{hilton2022Batch} originally showed, and we confirmed, this is largely due to the implicitly stronger regularisation at higher numbers of parallel environments. One remedy for this is to use PPO-EWMA, with a fixed COM, but a large number of parallel environments---leading to a consistent level of regularisation while achieving fast wall-clock times.

\section{Limitations \& Future work}
\looseness=-1 As with any conceptual model, there are simplifications and situations where the model may not apply. While we have shown that our model makes accurate predictions in certain settings, future work should investigate where this model breaks down, and where additional care is required. In particular, our analysis focuses solely on dense reward tasks, which can be viewed similarly to smooth optimisation problems, where small changes in the input lead to small changes in the output. 
Future work is required to more deeply understand sparse-reward tasks and environments that rely on strong exploration.
Additionally, scaling parallelisation can be cost-prohibitive or impossible in many settings. However, one can perfectly simulate a larger number of parallel environments by accumulating experience into a buffer before training on it, although this requires additional storage, and may be too slow for practical purposes.
This is why we performed our final experiments on Kinetix, as its hardware-accelerated implementation allows us to collect trillions of transitions within a reasonable wall clock time. In addition, prior baselines in Kinetix were far from optimal, so there is a lot of headroom to show that our techniques can be helpful, which is not the case in many other benchmarks. In \cref{app:isaacgym} we have preliminary results in Isaacgym, showing that increasing parallelisation leads to higher asymptotic performance; however, future work should thoroughly investigate how the insights from our work transfer to other domains. In this vein, \citet{devvrit2026theArt} investigate scaling LLM RL, also finding that larger batch sizes lead to higher asymptotic performance, suggesting that our results may have wide applicability.
Finally, using a fixed batch size throughout training may not be optimal, and investigating adaptive batch size techniques is another promising avenue for future work; see \citet{park2026scalable} for an interesting piece of work in this direction.

\section{Conclusion}
While there are many causes of plateaus in RL, in this work we demonstrate that under-regularisation is one reason behind learning stagnation in deep on-policy algorithms. 
By comparing PPO to stochastic optimisation, we find that many of the pathologies in the latter setting can occur in the former setting if the outer step size is too large.
However, this is easy to remedy, by either increasing the regularisation (e.g., decreasing $\epsilon$ or increasing the proximal policy's COM in PPO-EWMA) or directly increasing the parallelisation factor.
Based on our insights, we recommend a simple approach to scaling PPO---keep the minibatch size fixed as the number of parallel environments is changed---which allows it to remain competitive with more complex methods, and reliably scale to more than 1M parallel environments.
Finally, we demonstrate that, in a difficult and open-ended setting where current approaches are far from optimal, simply increasing the parallelisation leads to performance increasing monotonically across orders of magnitudes more experience.
Ultimately, we hope our work is one step in the direction of designing RL algorithms that can predictably scale with additional compute, and continue to benefit from additional experience indefinitely.

\section*{Acknowledgements}
Thank you to Charlie Cowen-Breen and Vincent Roulet for helpful discussions throughout the course of this project. Part of the compute for this work was provided by the Isambard-AI National AI Research Resource, under the project ``FLAIR 2025 Moonshot Projects''.
MB is funded by the Rhodes Trust. JF is partially funded by the UKRI grant EP/Y028481/1 (originally selected for funding by the ERC),
the JPMC Research Award and the Amazon Research Award.

\bibliography{main}
\bibliographystyle{assets/rlj}
\newpage
\section*{Appendix}
\appendix

\section{Experimental Details}\label{app:env_details}
For all of our investigative experiments, we use the \texttt{Jax2D} physics engine~\citep{matthews2024kinetix}, where we construct a set of 512 procedurally-generated 2D robotic locomotion tasks. 
The goal in each task is to move the morphology to the goal position, marked by a blue circle, and we measure performance by calculating average success rate over these tasks, in line with \citet{matthews2024kinetix}.
Furthermore, we use Stoix's implementation of PPO~\citep{toledo2024stoix} for these experiments.
All experiments used NVIDIA A100 40GB GPUs.

\subsection{Hyperparameters}\label{app:hyperparameters}
For all experiments, we run five independent seeds, and plot the mean solve rate or return, with the 95\% CI shaded. For the Kinetix SFL experiments, we run three seeds due to computational constraints.

\textbf{Locomotion Tasks}
For all of the locomotion experiments, the hyperparameters stayed mostly the same, with the exception of the hyperparameters we sweep over for a particular plot. The default settings are given in \cref{table:ppo-hyperparams}.
We use a 3-layer MLP with width 256.

\textbf{SFL}
The PPO hyperparameters are given in \cref{table:ppo-hyperparams}, and the SFL environment filtering parameters are shown in \cref{table:sfl_hypers}. We use the same exact values as \citet{matthews2024kinetix} for the 2048 parallel environment run, but adjust these hyperparameters as we use additional hardware. In particular, since filtering is trivially parallelizable, we increase the number of levels we search through as we increase the number of GPUs without a noticeable wall-clock-time penalty.

\textbf{SAPG}
We use the same code and hyperparameters as \citet{singla2024sapg} do (see \cref{tab:sapg_hypers}), with the exception of the minibatch size. We run for 10B timesteps.

\textbf{Stochastic Optimisation Example}\label{app:sgd_example}
The stochastic optimisation problem we consider is to minimise $\textbf{x}^T\textbf{x}$, with $\textbf{x} \in \mathbb{R}^{50}$. We perform standard gradient descent, but add noise with standard deviation $\frac{3}{\sqrt{50}}$ to the gradients. To more clearly demonstrate the similarities to RL, we plot the negative of the euclidean distance to the optimal solution $\textbf{x}^* = \textbf{0}$.

\begin{table}[h]
    \caption{Hyperparameters for the investigative and large-scale open-ended learning experiments.}
    \label{table:ppo-hyperparams}
    \begin{center}
    \scalebox{1.0}{
        \begin{tabular}{lccc}
        \toprule
        \textbf{Parameter} & \textbf{Jax2D Locomotion}  & \textbf{SFL} \\
        \midrule
        $\gamma$                        & 0.995  & 0.995 \\
        $\lambda_{\text{GAE}}$          & 0.9    & 0.9   \\
        PPO number of steps             & 256      & 256   \\
        PPO epochs                      & 8          & 8     \\
        PPO $\epsilon$                  & 0.2    & 0.2  \\
        PPO max gradient norm           & 0.5      & 0.5   \\
        PPO value clipping              & yes     & yes   \\
        Value loss coefficient          & 0.5      & 0.5   \\
        Entropy coefficient             & 0.01   & 0.01  \\
        PPO \# parallel environments    & 2048  & ---  \\
        Adam learning rate              & 0.0003    & 5e-5  \\
        PPO minibatches per epoch       & 32   & ---  \\
        \bottomrule 
        \end{tabular}}
    \end{center}
\end{table}

\begin{table}[h]
    \caption{Configuration for the different SFL~\citep{rutherford2024noRegrets} runs. $L$ is the rollout length used to compute the learnability score per environment, $\rho$ is the fraction of high-learnability environments used (the rest being filled with random levels), $N$ is how many levels we search through and $K$ is how many levels we save. Finally, $T$ is the number of PPO update steps between buffer updates, where each iteration consists of $256 \cdot N_\text{envs}$ transitions. We set $T$ such that we have the same number of environment transitions between buffer update steps for the 8k/65k/1M settings.
    }
    \label{table:sfl_hypers}
    \begin{center}
    \scalebox{1.0}{
        \begin{tabular}{lcccc}
        \toprule
        \textbf{Parameter} & \textbf{2048} & \textbf{8192} & \textbf{65536} & \textbf{1M} \\
        \midrule
        Rollout Length $L$      & 512   & 512   & 512   & 512   \\
        Sample Ratio $\rho$     & 0.5   & 0.5   & 0.5   & 0.5   \\
        Filtering Batch Size $N$          & 12288  & 256k  & 256k  & 4M  \\
        Update Period $T$       & 128  & 256  & 32  & 2  \\
        Buffer Size $K$         & 1024  & 8192  & 8192  & 8192  \\
        \bottomrule 
        \end{tabular}}
    \end{center}
\end{table}

\begin{table}[h]
    \centering
    \caption{Training Hyperparameters for AllegroKuka, Shadow Hand, and Allegro Hand. The values here were taken directly from \citet{singla2024sapg} and our results were obtained using the code \href{https://github.com/jayeshs999/sapg}{here}.}
    \label{tab:sapg_hypers}
    \begin{tabular}{lccc}
        \toprule
        \textbf{Hyperparameter} & \textbf{AllegroKuka} & \textbf{Shadow Hand} & \textbf{Allegro Hand} \\
        \midrule
        Discount factor, $\gamma$    & 0.99           & 0.99           & 0.99           \\
        $\tau$                       & 0.95           & 0.95           & 0.95           \\
        Learning rate                & 1e-4           & 5e-4           & 5e-4           \\
        KL threshold for LR update   & 0.016          & 0.016          & 0.016          \\
        Grad norm                    & 1.0            & 1.0            & 1.0            \\
        Entropy coefficient          & 0              & 0              & 0              \\ 
        Clipping factor $\epsilon$   & 0.1            & 0.1            & 0.2            \\
        Critic coefficient $\lambda'$ & 4.0            & 4.0            & 4.0            \\
        Horizon length               & 16             & 8              & 8              \\
        LSTM Sequence length         & 16             & ---            & ---            \\
        Bounds loss coefficient      & 0.0001         & 0.0001         & 0.0001         \\
        Mini epochs                  & 2              & 5              & 5              \\
        \bottomrule
    \end{tabular}
\end{table}

\FloatBarrier

\section{Center of Mass vs $\epsilon$}\label{app:beta_vs_epsilon}
\begin{figure}[H]
    \begin{subfigure}[t]{0.62\linewidth}    
        \centering
        \includegraphics[width=1\linewidth]{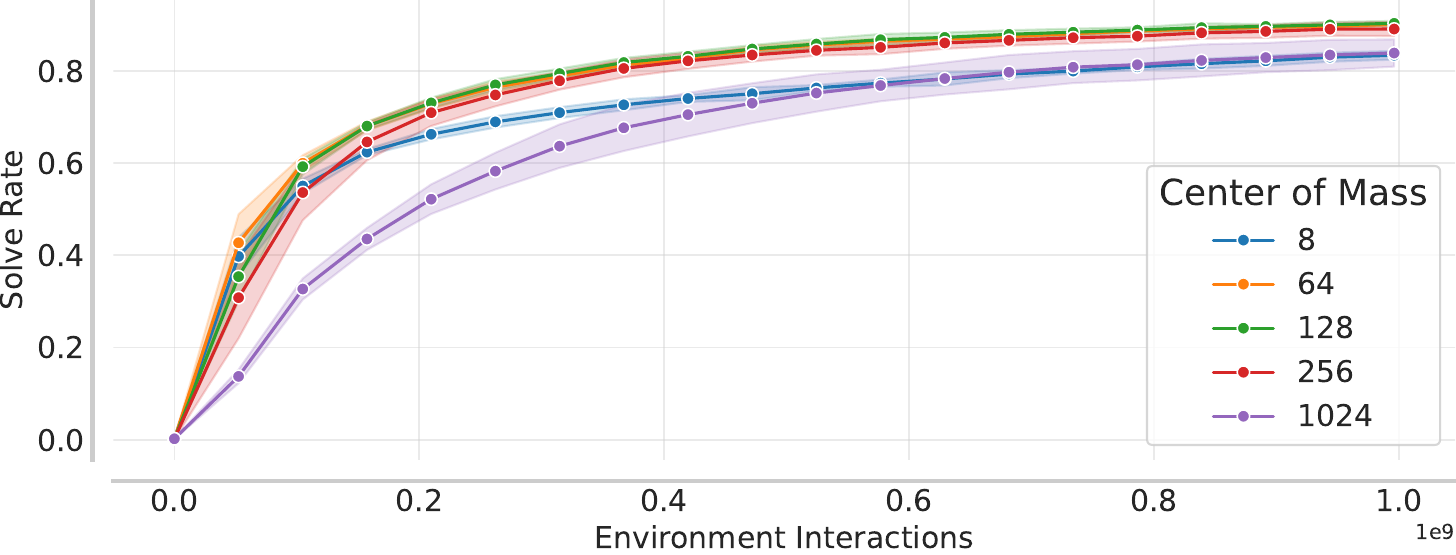}
        \caption{}
        \label{fig:best_beta_for_each_epsilon}
    \end{subfigure}%
    \begin{subfigure}[t]{0.38\linewidth}    
        \centering
        \includegraphics[width=1\linewidth]{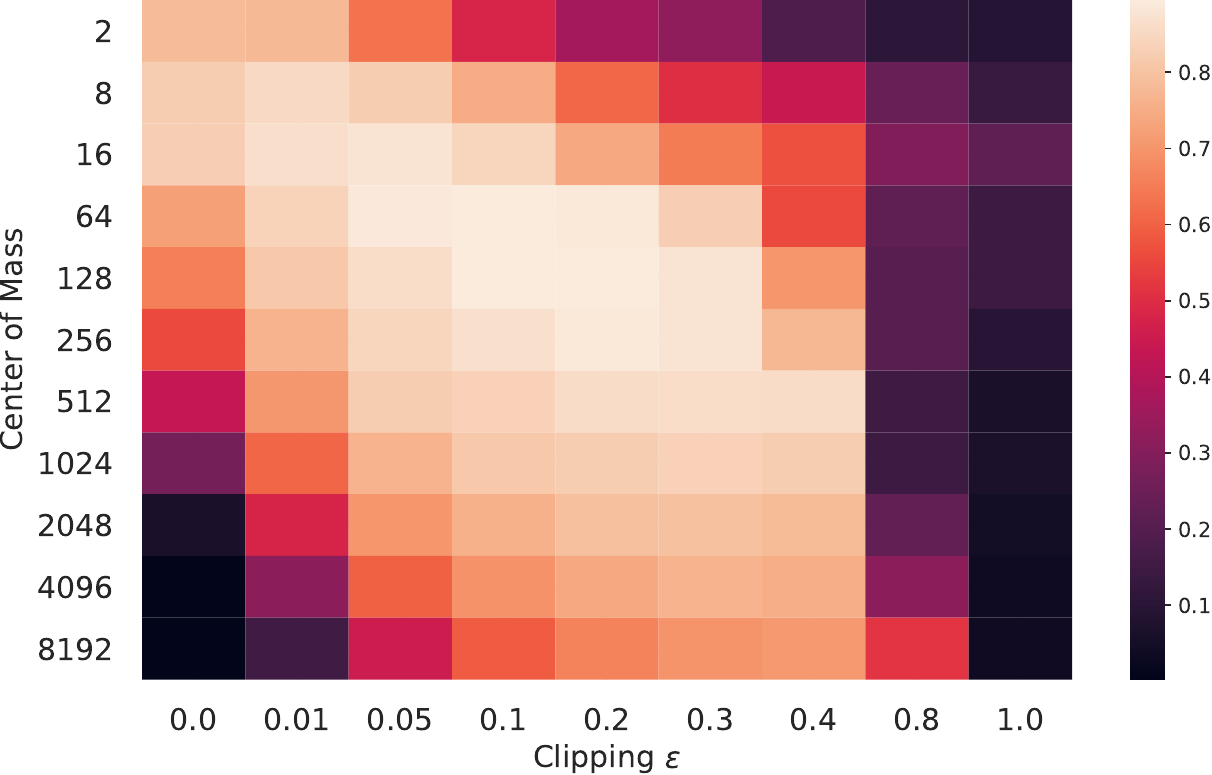}
        \caption{}
        \label{fig:beta_epsilon_heatmap}
    \end{subfigure}
    \caption{Comparing the effect of COM vs $\epsilon$. (a) Showing the performance of the \textit{best} $\epsilon$ for various COMs, showing that we can find an $\epsilon$ to (mostly) counteract the effect of changing the COM in PPO-EWMA. (b) A heatmap of final performance for a 2D grid search over the PPO-EWMA COM and $\epsilon$. Overall, most reasonable values of the COM have a corresponding $\epsilon$ that performs well; however, extreme values of $\epsilon$ are too unstable to learn.}\label{fig:comparing_beta_and_eps}
\end{figure}

In \cref{fig:comparing_beta_and_eps}, we compare the effect of changing the COM vs changing the clipping $\epsilon$. 
One takeaway from this plot is that we can mostly counteract changes in one of these quantities by appropriately altering the other, suggesting that both of these settings act on the same mechanism.

However, one difference is in the susceptibility of the agent to $\epsilon$-overshooting. As mentioned in \cref{sec:modulates_epochs}, $\epsilon$ does not actually constrain the ratio to be within the $1 \pm \epsilon$ range; instead, it stops updates once the ratio already exceeds the bounds. Therefore, if the Adam learning rate is high enough, there is little difference between all $\epsilon$ values lower than some threshold, which is roughly how much one gradient step can change the probability ratio of a particular action. 
However, when increasing the COM of PPO-EWMA, this overshooting is less of a problem, since we are measuring the ratio with respect to the (potentially quite old) proximal policy, meaning that the gradients are only non-zero when the proximal policy has caught up enough with the current behaviour policy, i.e., when the ratio is within the $1 \pm \epsilon$ range.

Finally, we find that when studying very weak regularisation, a low COM in PPO-EWMA tends to be more stable than a high $\epsilon$ (e.g., compare the first row of \cref{fig:beta_epsilon_heatmap} with the last column). One potential reason for this is that with a large $\epsilon$, single, perhaps unreliable transitions can lead to large gradients which can drown out the actual signal. These large gradients can also potentially cause destructive weight updates that completely change the policy's behaviour. 

\section{Additional Scaling Results}\label{app:additional_scaling_results}

\begin{figure}[htbp]
  \centering
  \begin{minipage}[t]{0.48\textwidth}
    \centering
    \includegraphics[width=\linewidth]{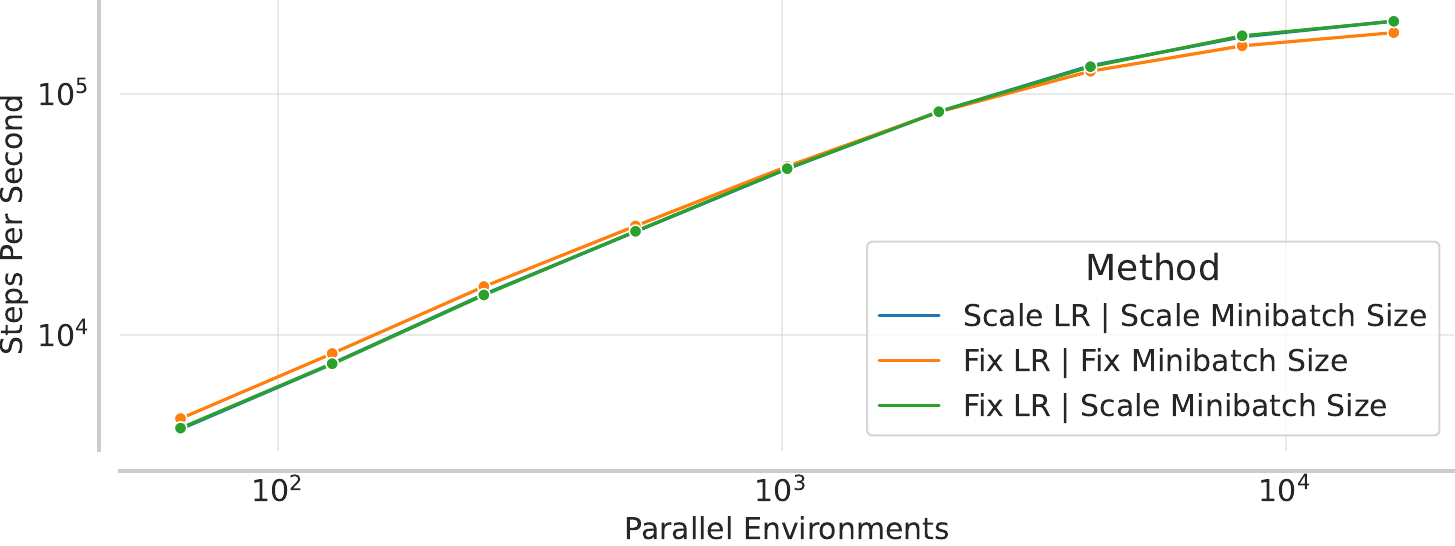}
    \caption{Plotting the steps per second for the training, which includes the environment step and the neural network optimisation. This corresponds to the same results as in \cref{fig:how_to_scale_ppo}.}
    \label{fig:ppo_scaling_sps}
  \end{minipage}
  \hfill 
  \begin{minipage}[t]{0.48\textwidth}
    \centering
    \includegraphics[width=\linewidth]{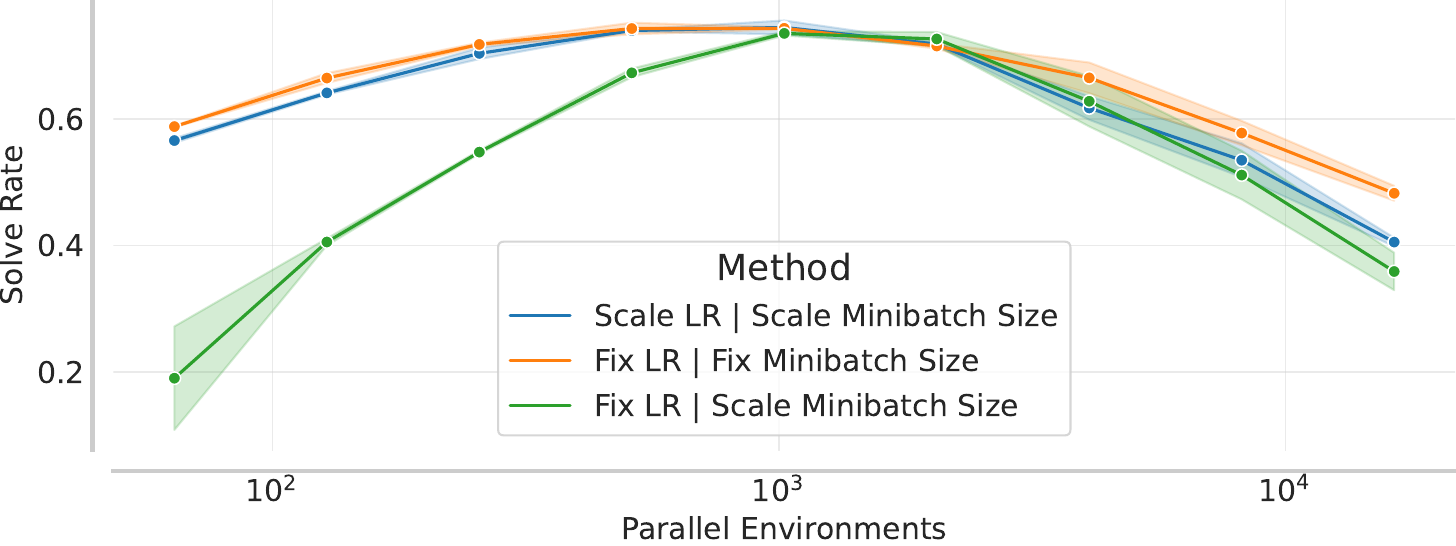}
    \caption{Comparing different approaches when changing the number of parallel environments. }
    \label{fig:how_to_scale_ppo_summary}
  \end{minipage}
\end{figure}

\begin{figure}[h]
    \centering
    \includegraphics[width=1\linewidth]{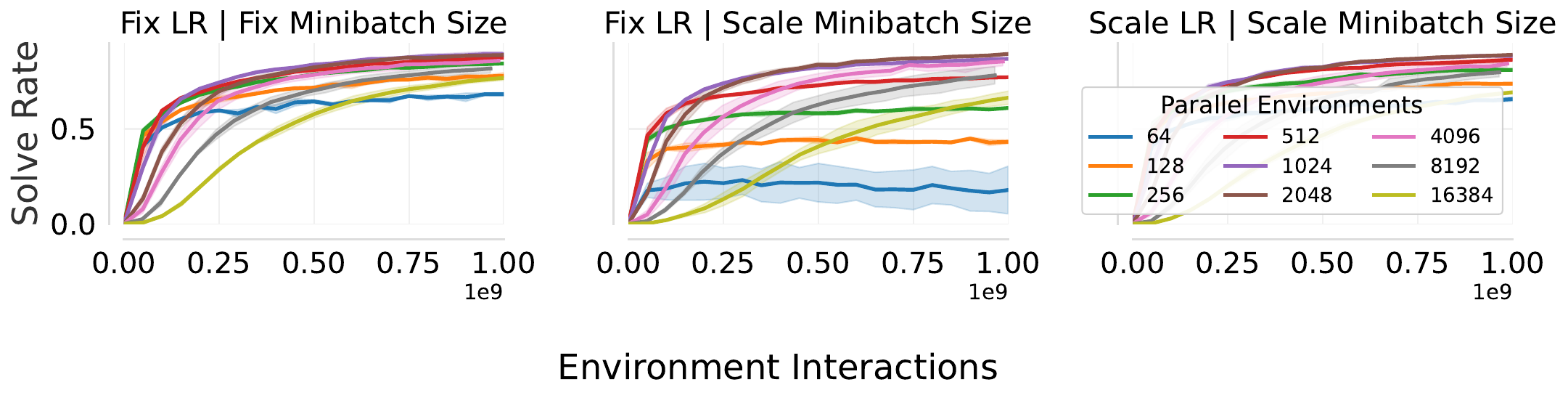}
    \caption{A version of \cref{fig:how_to_scale_ppo} with more options for the number of parallel environments.}
    \label{fig:how_to_scale_ppo_full_lines_appendix}
\end{figure}

\cref{fig:ppo_scaling_sps} shows the number of environment steps per second we can process in the locomotion task, when taking into account all learning and environment stepping. We see that there is little difference between the various scaling approaches, and this translates to little difference in overall wall-clock time. \cref{fig:how_to_scale_ppo_summary} provides a condensed version of \cref{fig:how_to_scale_ppo}, whereas \cref{fig:how_to_scale_ppo_full_lines_appendix} contains more granular data.

\FloatBarrier
\section{SFL Scaling}\label{app:sfl_scaling}
\begin{figure}[h]
    \centering
    \includegraphics[width=1\linewidth]{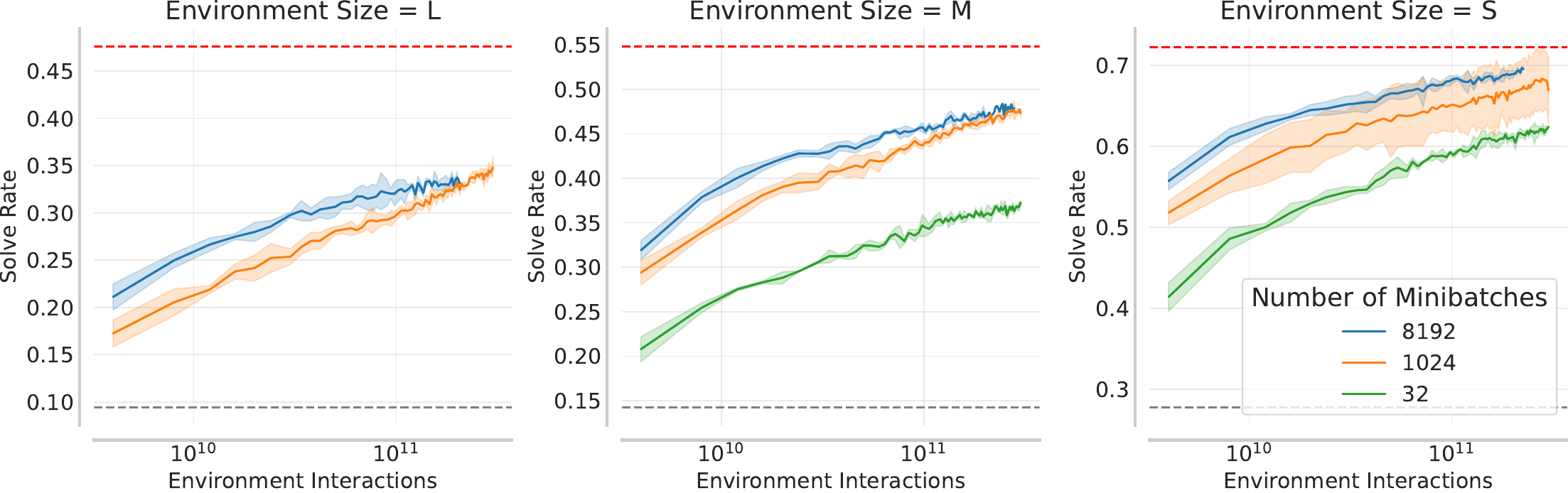}
    \caption{Comparing performance when using 1M parallel environments, but a different number of minibatches. \citet{matthews2024kinetix} use 32 minibatches for the default setting of 2048 parallel environments, and using this for 1M parallel environments performs significantly worse, in line with the results from \cref{fig:how_to_scale_ppo}. The \texttt{large} environment size runs out of memory when using only 32 minibatches.}
    \label{fig:sfl_ablations_1M_minibatches}
\end{figure}

\begin{figure}[h]
    \centering
    \begin{subfigure}[b]{0.8\linewidth}
        \centering
        \includegraphics[width=\linewidth]{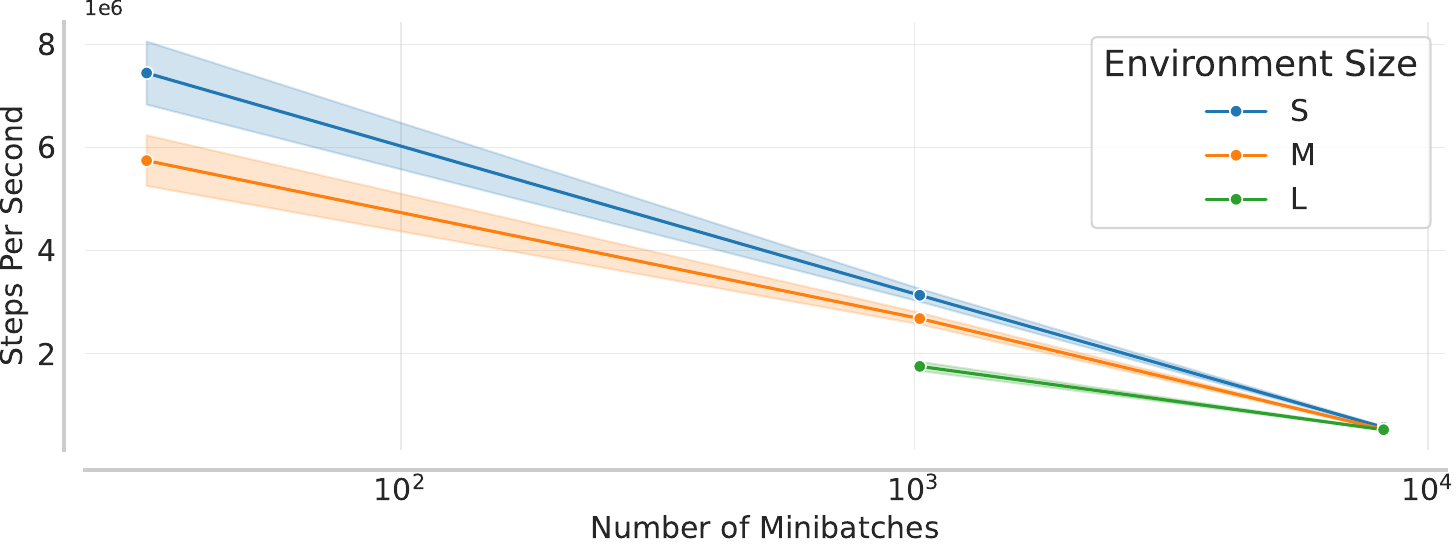}
        \caption{Steps per second vs. minibatches.}
        \label{fig:sfl_ablations_1M_minibatches_sps_per_minibatch}
    \end{subfigure}
    \\
    \begin{subfigure}[b]{0.8\linewidth}
        \centering
        \includegraphics[width=\linewidth]{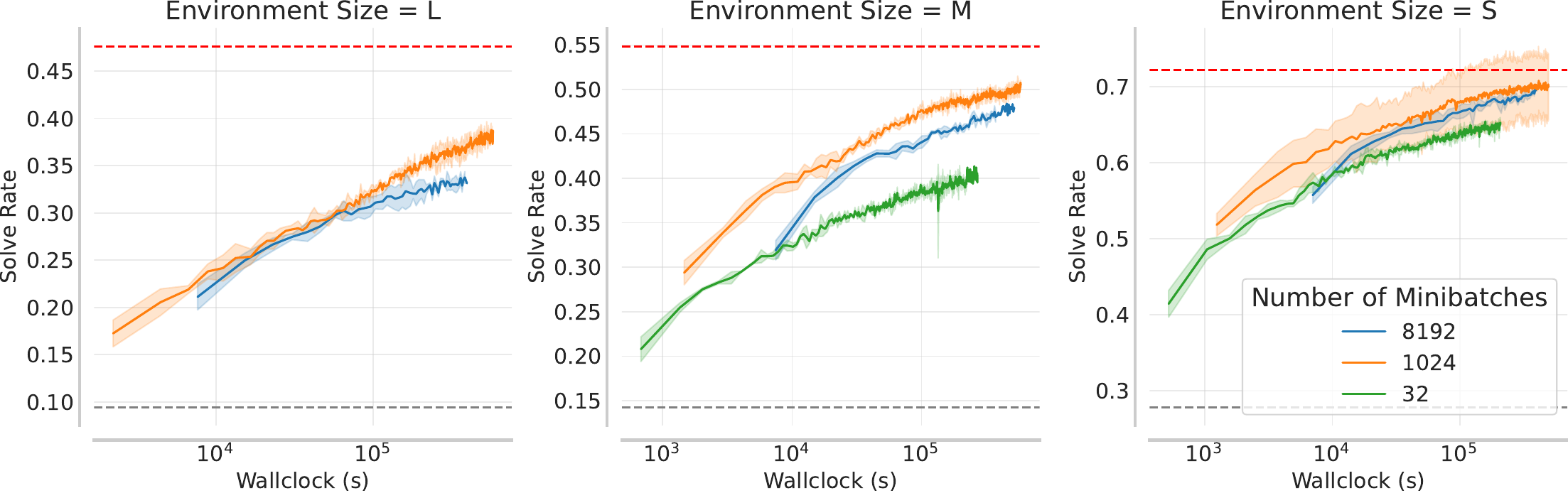}
        \caption{Performance as a function of wall-clock time.}
        \label{fig:sfl_ablations_1M_minibatches_wallclock}
    \end{subfigure}

    \caption{Comparing runtime when using different minibatch sizes and 1M parallel environments. Note that \texttt{L} runs out of memory with 32 minibatches.}
    \label{fig:sfl_ablations_combined}
\end{figure}

For the primary SFL results, we scaled to $1048576$ parallel environments, which is $512 \times $ more than the default used by \citet{matthews2024kinetix}. 
According to our recipe, this means we must have $512 \times $ the number of minibatches, i.e., 16384 instead of the default 32. However, we parallelise across 128 GPUs, meaning each GPU has 8192 parallel environments, and due to how the baseline code is written (which we wanted to keep unchanged), we cannot have more minibatches than we have parallel environments. 
In other words, we must have 8192 or fewer minibatches. For our main results, we use 1024 minibatches ($32\times$ more than the default), meaning each minibatch is $16\times$ larger than the default; we therefore scale the learning rate by $4 = \sqrt{16}$ to account for this.
This setting provides a balance between performance and wall-clock time, and the difference in performance between the 1024 minibatch setting and the 8192 minibatch setting reduces as we train for longer, as evidenced by \cref{fig:sfl_ablations_1M_minibatches}. Furthermore, the wall-clock time difference between 8192 minibatches and 1024 minibatches is substantial, since the former case does not come close to saturating the GPUs (see \cref{fig:sfl_ablations_1M_minibatches_sps_per_minibatch}).
Taken together, in \cref{fig:sfl_ablations_1M_minibatches_wallclock}, we see that the 1024 minibatch setting provides the best performance as a function of wallclock-time, and is not materially different to the final performance of the 8192 minibatch setting, even if the latter is slightly more sample efficient.

\FloatBarrier
\section{SFL Ablations}\label{app:sfl_ablations}
In this section we perform some ablations on the SFL results, to demonstrate which factors influence performance. For all cases, the ablations and ``baseline'' use 8192 environments, so that we could train for more timesteps than if we had used 2048.

\subsection{Learning Rate}\label{app:sfl_ablations_lr}

\begin{figure}[H]
    \centering
    \includegraphics[width=1\linewidth]{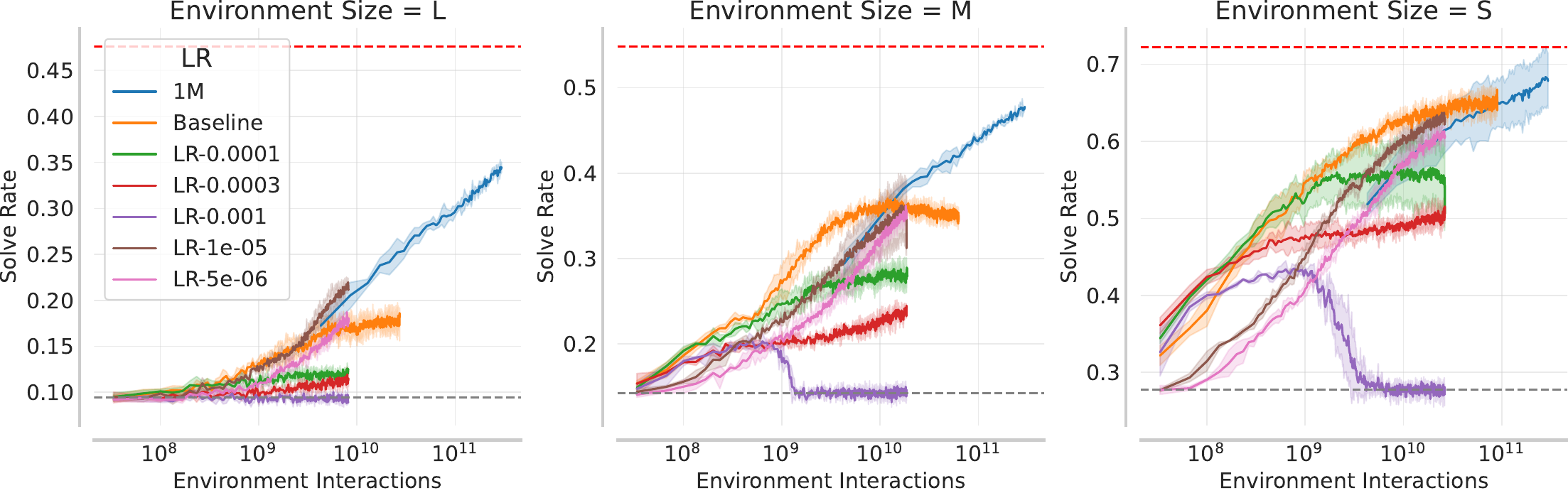}
    \caption{SFL results when keeping the number of environments fixed at 8192, but changing the learning rate. The baseline ($5e-5$) and 1M parallel environment run are shown for reference. Reducing the learning rate can avoid plateaus, but training is too slow to be able to process enough environment samples within a reasonable time.}
    \label{fig:sfl_ablations_lr}
\end{figure}

\cref{fig:sfl_ablations_lr} shows the effect of changing the learning rate. While reducing the learning rate by a factor of $5\times$ to $1e-5$ can avoid the early plateauing, it requires a prohibitively long wall-clock time to obtain a large amount of samples, whereas increasing parallelisation allows us to obtain the result in significantly less time, albeit at the cost of additional hardware.

\subsection{PPO-EWMA}\label{app:sfl_ablations_ewma}
\begin{figure}[H]
    \centering
    \includegraphics[width=1\linewidth]{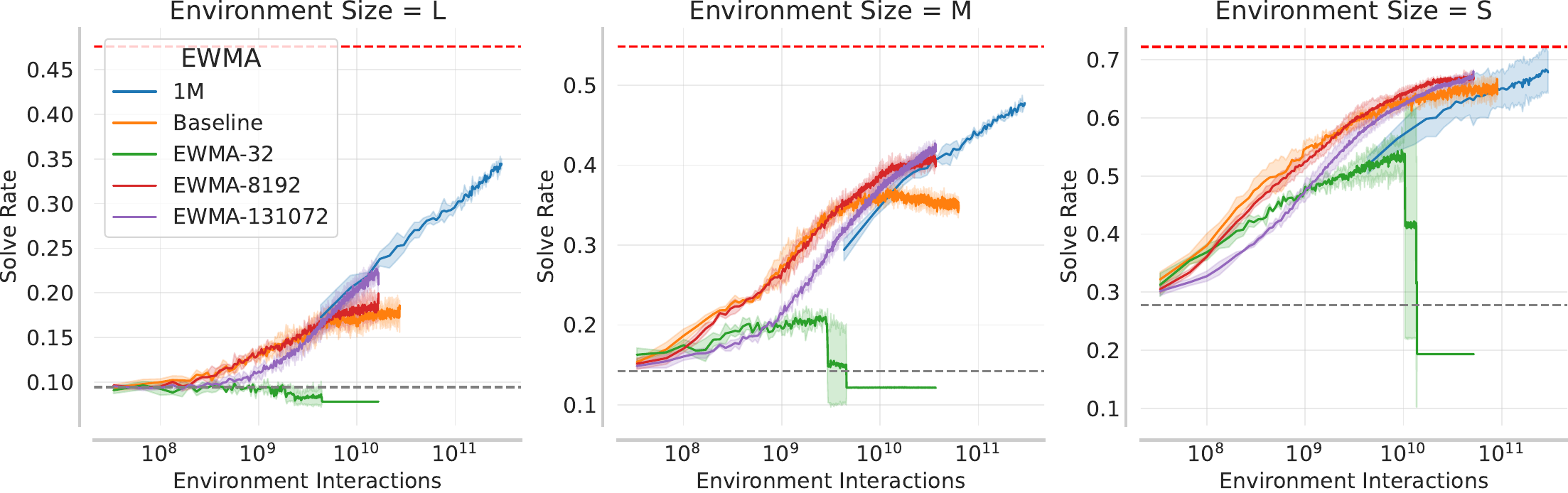}
    \caption{SFL results when keeping the number of environments fixed at 8192, but using PPO-EWMA. The baseline and 1M parallel environment run, both using normal PPO, are shown for reference. Increasing the center of mass, and thereby the regularisation, can alleviate plateaus.}
    \label{fig:sfl_ablations_ewma}
\end{figure}

Next, in \cref{fig:sfl_ablations_ewma}, we show that using PPO-EWMA with a large center of mass can also improve performance over the baseline, and plateaus later, supporting the argument that the smaller outer step size due to increased parallelisation is beneficial.

\subsection{Additional Filtering}\label{app:sfl_ablations_filtering}
\begin{figure}[H]
    \centering
    \includegraphics[width=1\linewidth]{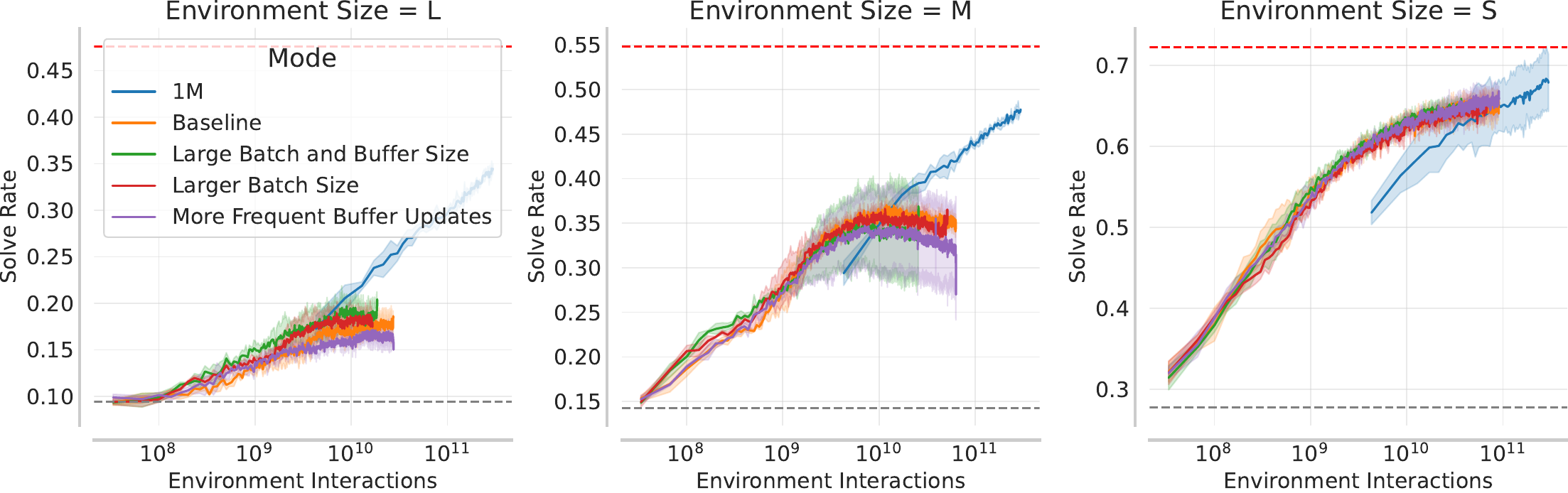}
    \caption{SFL results using 8192 environments, but either sampling more levels to filter through, or doing this and having a larger buffer of stored, high-learnability levels. This shows that while the 1M parallel environment run samples more levels, and has a larger buffer, these changes are insufficient to prevent the plateau that the 8192 environments agent succumbs to.}
    \label{fig:sfl_ablations_sampling}
\end{figure}

Finally, the 1M parallel environment run performed additional filtering of environments to select the subset we actually train on. 
In particular, each GPU processed the same number of levels, meaning that the total number of pre-filtering levels we considered is larger than the baseline. We further increased the size of the high-learnability level buffer we sample from, since we have orders of magnitude more parallel environments.
In \cref{fig:sfl_ablations_sampling}, we show that these changes alone are insufficient to improve performance, unless they are also coupled with increasing the parallelisation.

\section{SFL Hand Designed Results}\label{app:sfl_hand_designed_results}

\begin{figure}[H]
    \centering
    \includegraphics[width=1\linewidth]{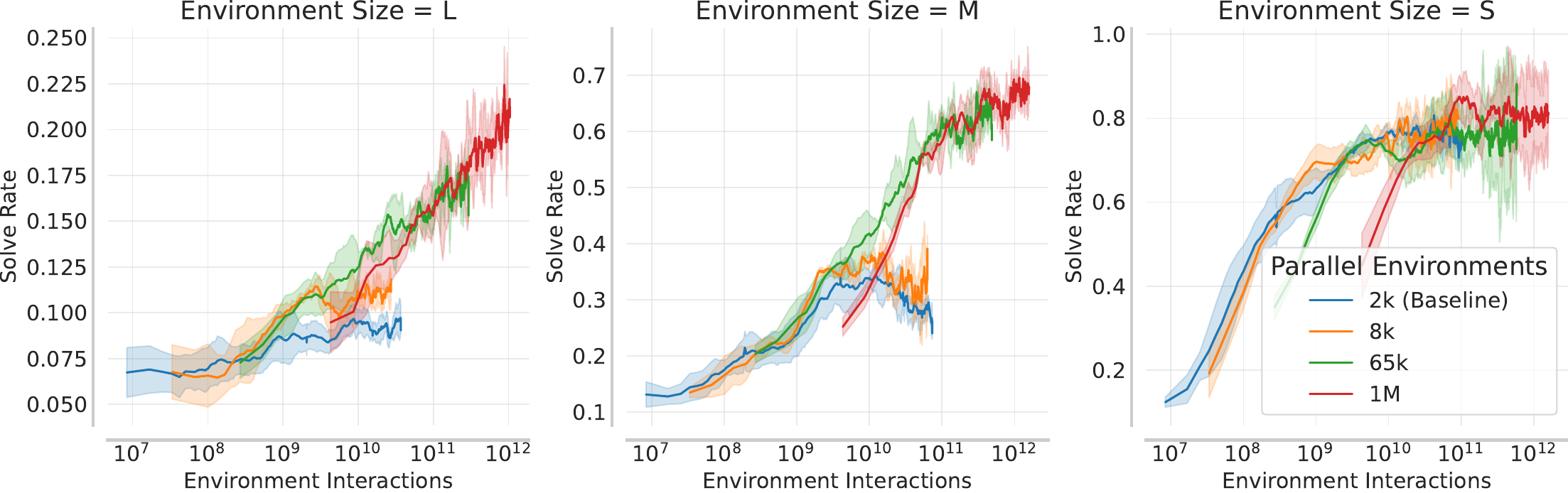}
    \caption{SFL Results on hand-designed environments. While the performance is much more noisy, the same rough trend holds, where additional parallelisation improves performance.}
    \label{fig:sfl_hand_designed}
\end{figure}

\cref{fig:sfl_hand_designed} contains the same agents as in \cref{fig:sfl_main_results}, but measured on the hand-designed evaluation set of levels~\citep{matthews2024kinetix}. The same overall trend is visible, in that more parallel environments lead to higher asymptotic performance.

\section{Isaacgym Parallelisation Results}\label{app:isaacgym}
\begin{figure}[H]
    \centering
    \includegraphics[width=1\linewidth]{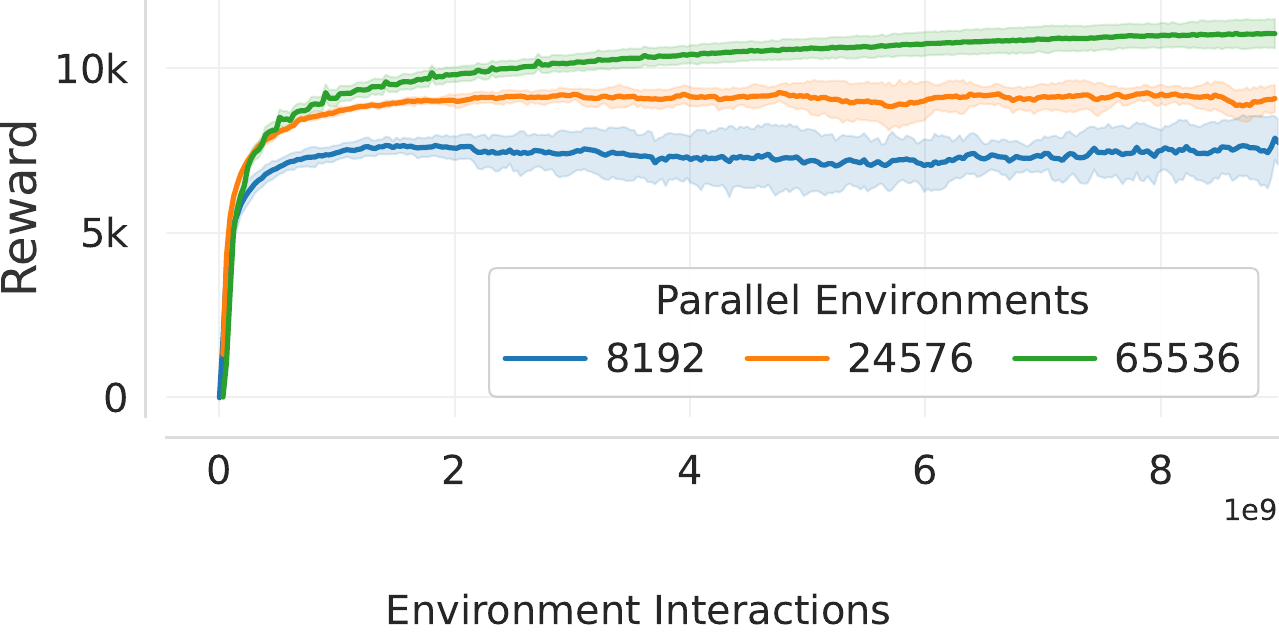}
    \caption{Scaling the number of parallel environments in the Isaacgym task Shadow Hand.}
    \label{fig:sapg_numenvs_sweep}
\end{figure}

\cref{fig:sapg_numenvs_sweep} shows that increasing parallelisation leads to higher asymptotic performance in Isaacgym. 

\end{document}